\documentclass[10pt,twocolumn,letterpaper]{article}

\usepackage[pagenumbers]{cvpr} 

\definecolor{cvprblue}{rgb}{0.21,0.49,0.74}
\usepackage[pagebackref,breaklinks,colorlinks,allcolors=cvprblue]{hyperref}

\usepackage{booktabs}
\usepackage{multirow}
\usepackage{graphicx}
\usepackage{amsmath}
\usepackage{booktabs}
\usepackage[normalem]{ulem}
\usepackage[accsupp]{axessibility}  

\newcommand{\myparagraph}[1]{\smallskip\noindent\textbf{#1}}

\newcommand{\ours}{HistoSelect\xspace}

\def\paperID{} 
\def\confName{CVPR}
\def\confYear{2026}

\title{Act Like a Pathologist: Tissue-Aware Whole Slide Image Reasoning}

\author{
Wentao Huang\textsuperscript{1,2}\thanks{Email: Wentao Huang (wenthuang@cs.stonybrook.edu). \\ \hphantom{***} Work done during internship at Mayo Clinic.}\hskip 1em
Weimin Lyu\textsuperscript{1}\hskip 1em
Peiliang Lou\textsuperscript{2}\hskip 1em
Qingqiao Hu\textsuperscript{1}\hskip 1em
Xiaoling Hu\textsuperscript{3}\hskip 1em \\
Shahira Abousamra\textsuperscript{4}\hskip 1em
Wenchao Han\textsuperscript{2}\hskip 1em
Ruifeng Guo\textsuperscript{2}\hskip 1em
Jiawei Zhou\textsuperscript{1}\hskip 1em
Chao Chen\textsuperscript{1}\hskip 1em
Chen Wang\textsuperscript{2}\hskip 1em
\\
\textsuperscript{1}Stony Brook University, NY, USA \hskip 1em
\textsuperscript{2}Mayo Clinic, MN, USA \hskip 1em
\\
\textsuperscript{3}Harvard Medical School, MA, USA \hskip 1em
\textsuperscript{4}Stanford University, CA, USA \hskip 1em
}

\begin{document}
\maketitle

\begin{abstract}

Computational pathology has advanced rapidly in recent years, driven by domain-specific image encoders and growing interest in using vision–language models to answer natural-language questions about diseases. Yet, the core problem behind pathology question-answering remains unsolved, considering that a gigapixel slide contains far more information than necessary for a given question. Pathologists naturally navigate tissue and morphology complexity by scanning broadly, and zooming in selectively according to the clinical questions. Current models, in contrast, rely on uniform patch sampling or broad attention maps, often attending equally to irrelevant regions while overlooking key visual evidence. In this work, we try to bring models closer to how humans actually examine slides. We propose a question-guided, tissue-aware, and coarse-to-fine retrieval framework, HistoSelect, that consists of two key components: a group sampler that identifies question-relevant tissue regions, followed by a patch selector that retrieves the most informative patches within those regions.
By selecting only the most informative patches, our method becomes significantly more efficient: reducing visual token usage by 70\% on average, while improving accuracy across three pathology QA tasks. Evaluated on 356,000 question–answer pairs, our approach outperforms existing methods and produces answers grounded in interpretable, pathologist-consistent regions. Our results suggest that bringing human-like search and attention patterns into WSI reasoning is a promising direction for building practical and reliable pathology VLMs. Code is available at \url{https://github.com/winston52/HistoSelect}.

\end{abstract}

\section{Introduction}
\label{sec:introduction}

\begin{figure}[t]
\centering
\includegraphics[width=0.5\textwidth]{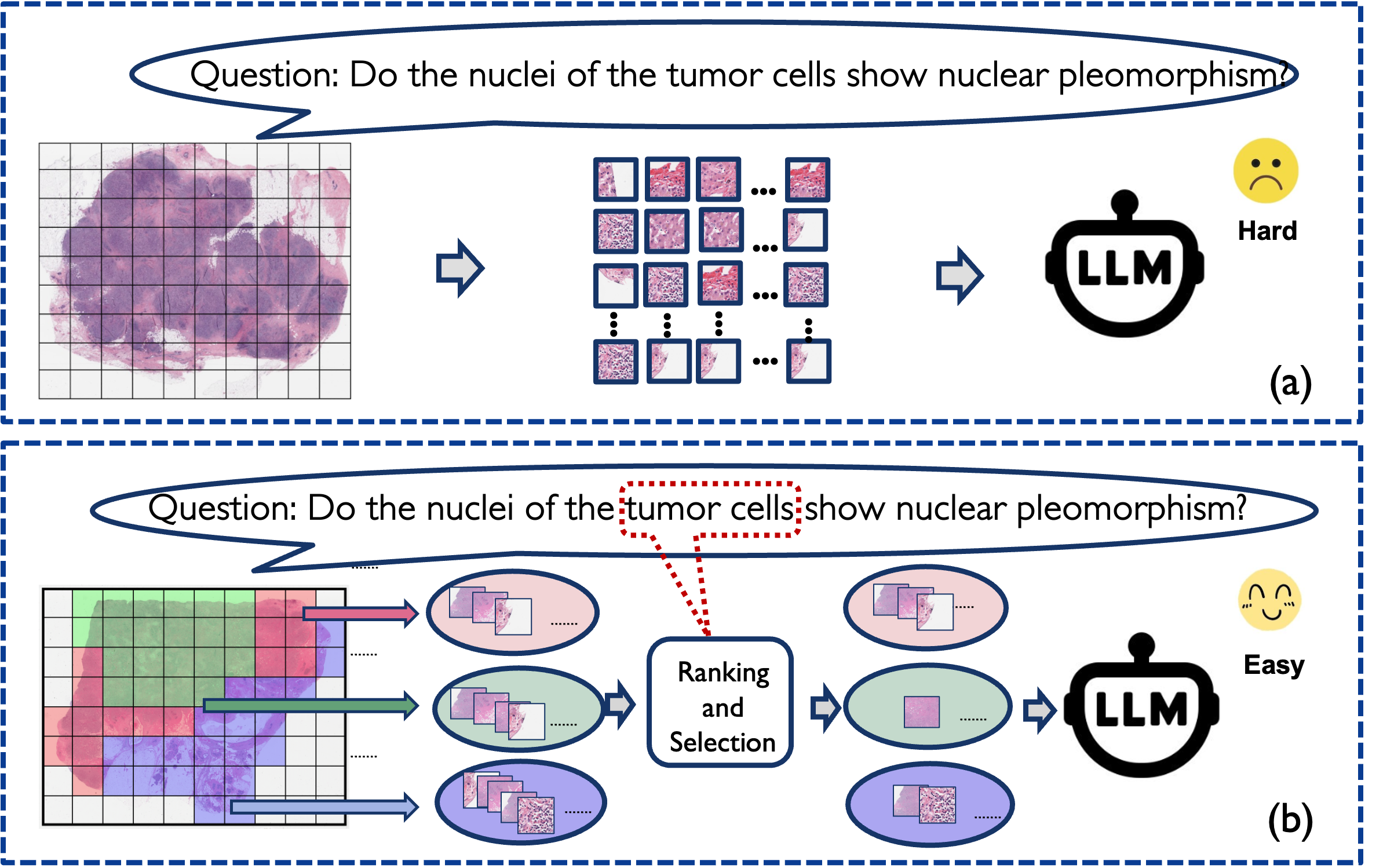}
\caption{Illustration of our HistoSelect framework.
    \textbf{(a)} The baseline method feeds a large number of patches indiscriminately into the VLM, leading to high redundancy and question-irrelevance. 
    \textbf{(b)} Our question-guided tissue-aware selection method. The question guides the model to select a relevant and sparse subset of informative patches, which are then fed to the VLM for reasoning.}
\label{fig:idea}
\end{figure}

\begin{figure*}[t]
\centering
\includegraphics[width=1.0\textwidth]{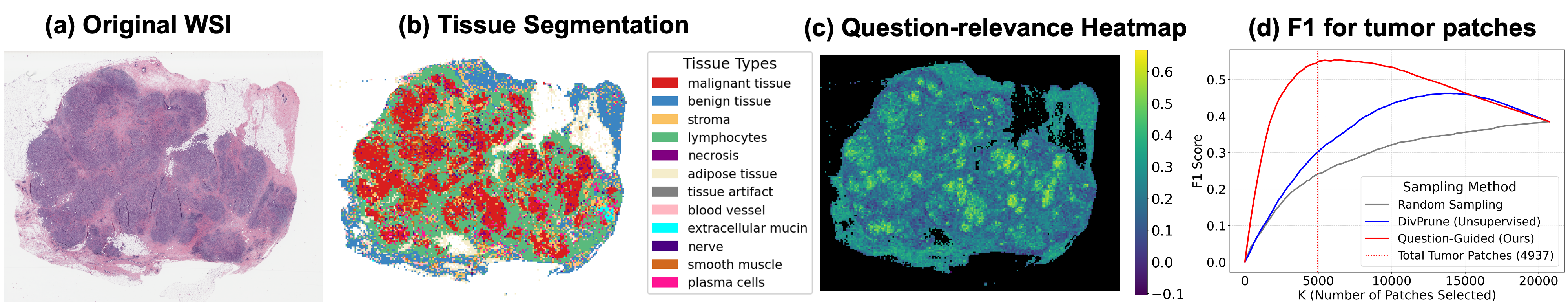}
\caption{Visualization and quantitative pre-analysis of patch relevance for a VQA sample (from TCGA-BRCA). (a) Reference WSI. (b) Tissue segmentation, with tumor region shown in red. (c) Patch-level relevance heatmap based on question-patch similarity. High-relevance regions (light region) align with the tumor region from (b). (d) F1 score comparison for retrieving tumor patches using different sampling methods. The question-guided (red) sampling strategy vastly outperforms question-agnostic methods like diversity sampling \cite{alvar2025divprune} (blue)  and random sampling (gray), demonstrating the limited efficacy of non-guided selection.}
\label{fig:motivation}
\end{figure*}

Histopathology image analysis plays a critical role in cancer diagnosis and treatment planning~\cite{lu2021ai,niazi2019digital,barisoni2020digital}. A key data source in this domain is the Whole Slide Image (WSI), a giga-pixel digital scan that captures rich cellular and tissue morphology.  
The rapid development of computational pathology has enabled success in fundamental tasks, such as subtype classification \cite{ilse2018attention, li2021dual, zhang2022dtfd, shao2021transmil, qu2023boosting, tang2023multiple, zhang2023attention, huang2024hard}, segmentation \cite{xu2025match, xu2024semi} and survival analysis \cite{chen2021multimodal, xu2023multimodal, zhou2023cross, zhang2024prototypical, song2024multimodal}. With the emergence of multi-modal learning, more challenging tasks have been introduced. In Visual Question Answering (VQA)~\cite{ikezogwo2023quilt, chen2024wsivqa, chen2025slidechat, liang2025wsi, sun2025pathbench}, models are required not only to predict a correct answer, but also to produce clinically trustworthy and interpretable answers.

To address the Pathology VQA task,  the classic Multiple Instance Learning (MIL) approaches \cite{chen2024wsivqa, ilse2018attention, li2021dual, lu2021data} are insufficient due to their lack of language understanding power.
Recent Multimodal Large Language Model (MLLM) based approaches~\cite{seyfioglu2024quilt, chen2025slidechat, liang2025wsi} convert WSI patches into visual tokens, which are then concatenated with the textual question tokens and fed into large language models (LLMs) for multi-modal reasoning. While these approaches have demonstrated competitive performance, they still suffer from two major limitations stemming from the nature of WSIs.  
The first challenge is the lack of attributable explainability. Although VQA models can generate textual answers, most existing MLLM methods do not reveal \emph{which} patches or regions in the WSI support the prediction.  This absence of localized, patch-level attribution results in a “black-box’’ behavior that undermines clinical trustworthiness, since pathologists cannot verify the model’s reasoning by inspecting the corresponding image evidence. The second challenge is the redundancy and question-irrelevance of patches. A single WSI can have tens of thousands of patches, many of which are irrelevant to the question, e.g., depicting background tissue, benign structures, or regions unrelated to the clinical decision. Meanwhile, patches from the same tissue type are often redundant; we do not need all of them to make the decision.
Furthermore, the strict token limits of current LLMs force existing methods to adopt question-agnostic strategies such as non-selective sampling~\cite{chen2025slidechat} or pooling~\cite{liang2025wsi}. This results in treating all patch tokens equally, which unnecessarily overwhelms the downstream LLM with question-irrelevant visual information and risks degrading the model's performance.

To address the aforementioned limitations, we take inspiration from how pathologists reason over WSIs. Rather than examining every region exhaustively, pathologists work in a tissue-aware manner: they first identify the tissue regions relevant to the clinical question and then zoom into a small set of critical patches for verification.
Following this principle, we first establish a coarse-grained tissue context.  
In collaboration with expert pathologists, we define a set of $K$ prompts describing fundamental tissue types, enabling a CLIP-like tissue segmentation that automatically assigns each WSI patch to a semantic category (\Cref{fig:motivation}b). This step mirrors the initial stage of locating diagnostically meaningful regions.
We further quantitatively validate fine-grained tissue region selection is guided by the question.  
By calculating the cosine similarity between each patch embedding and the question embedding, we generate a patch-level relevance heatmap (\Cref{fig:motivation}c), where lighter regions indicate high relevance to the tumor features specified by the question. 
A quantitative comparison (\Cref{fig:motivation}d) shows that question-guided sampling dramatically outperforms question-agnostic strategies (such as diversity sampling or random sampling) in retrieving relevant tumor patches. 
This validates our second key observation: the necessity of a question-guided selection mechanism to efficiently identify high-value information within gigapixel WSIs and lead to more accurate answers.

Motivated by these observations, we introduce \ours, a hierarchical, question-guided, and tissue-aware patch selection framework for the pathology VQA task, designed to mirror the coarse-to-fine diagnostic process of pathologists.  
As shown in \Cref{fig:idea}, we leverage the pathologist-defined prompts for fundamental tissue types and a pre-trained patch-level vision–language model~\cite{lu2024visual} to partition the WSI into semantically coherent tissue groups. This provides the coarse-grained structure upon which our method operates.
Building on this, \ours\ implements a two-stage selection mechanism grounded in the Information Bottleneck (IB) principle~\cite{tishby2000information}.  
The first stage, the \emph{group sampler}, evaluates how relevant each tissue type region (group) is to the input question and determines the patch sampling rate from each group.  
The second stage, the \emph{patch selector}, ranks the patches within each active group by relevance to the question and selects the most informative ones according to the allocated token budget. Together, these modules emulate the pathologist’s workflow: first identify the meaningful regions, then zoom in on the key evidence.

To ensure that the selected patches are both \emph{sparse} and \emph{sufficient} for answering the question, we formulate the training objective using a dual-level compression loss enforcing sparsity and relevance at both group- and patch-level.  
At both levels, we encourage the model to keep only what is necessary by penalizing the divergence between the learned selection probabilities and a dynamic prior derived from question–image similarity. 
This guides the selectors toward semantically aligned evidence while preventing over-selection.  
By integrating this IB-driven objective with the VQA loss, \ours\ produces a compact, question-aligned set of visual tokens that retains critical information and enhances interpretability.

In summary, our contributions are:

\begin{itemize}
\item We collaborate with pathologists to design a series of basic tissue type prompts, enabling us to partition the WSIs into distinct tissue regions.

\item We introduce HistoSelect, a hierarchical, question-guided, and tissue-aware selection framework based on the IB theory, which effectively prunes question-irrelevant tokens to increase the proportion of question-relevant tokens fed into the LLM for reasoning, thereby enhancing the model's interpretability.

\item We conduct a detailed pathologist evaluation to ensure that both our tissue segmentation and model selection results align with the expectations of clinical pathologists.

\item We achieve state-of-the-art performance on two public datasets and one in-house dataset.
\end{itemize}

\section{Related Work}
\label{sec:related_work}
\myparagraph{Whole Slide Image Analysis.} Traditional WSI analysis primarily focuses on slide-level classification \cite{ilse2018attention, li2021dual, zhang2022dtfd, shao2021transmil, qu2023boosting, tang2023multiple, zhang2023attention} and survival analysis \cite{chen2021multimodal, xu2023multimodal, zhou2023cross, zhang2024prototypical, song2024multimodal} using Multiple Instance Learning (MIL) \cite{ilse2018attention, li2021dual, zhang2022dtfd, qu2022bi, lin2023interventional, zhang20252dmamba, zhu2025dgr}. More recently, the field has advanced to Pathology Visual Question Answering (VQA) \cite{seyfioglu2024quilt,chen2025slidechat,liang2025wsi}, which is a more challenging task. Unlike the aggregation-focused objective of MIL, VQA demands fine-grained reasoning to answer queries ranging from global morphology to the identification of cellular features. Pathology VQA benchmarks include patch-level datasets, such as Quilt-LLaVA \cite{seyfioglu2024quilt}, and slide-level datasets like SlideChat \cite{chen2025slidechat} and WSI-LLaVA \cite{liang2025wsi}. In this work, we focus on developing a hierarchical selection method for the slide-level VQA task.

\begin{figure*}[t]
\centering
\includegraphics[width=.9\textwidth]{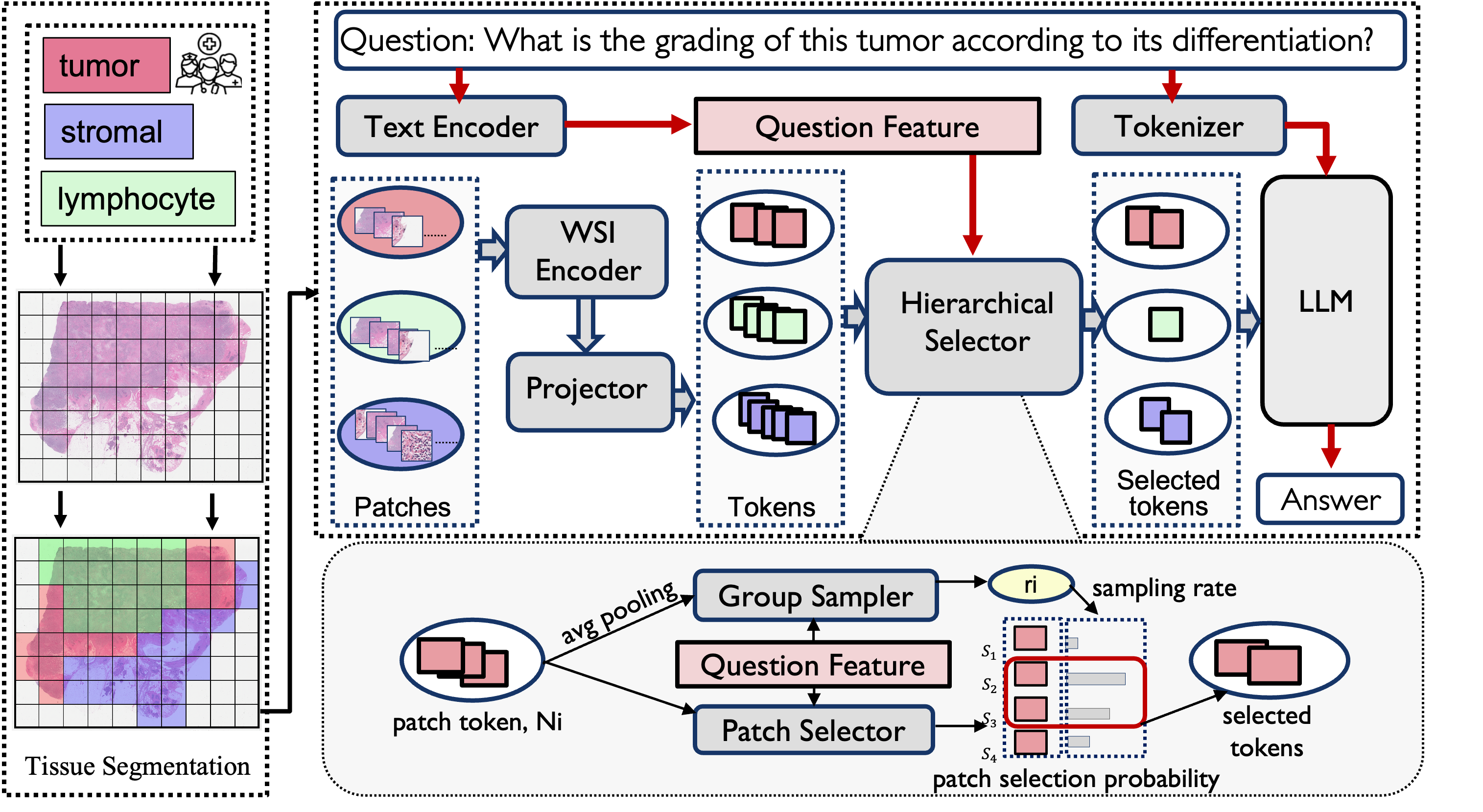}
\caption{Overview of HistoSelect. The framework operates in two stages: Tissue Segmentation partitions the WSI into $M$ tissue types (e.g., tumor, stromal, lymphocyte) using pathologist-designed prompts. The Hierarchical Selector then uses the question feature to dynamically select the top $K$ most relevant patch tokens, which are subsequently passed to the LLM for multi-modal answer generation.}
\label{fig:overview}
\end{figure*}

\myparagraph{Multi-Modal Histopathology Models.} Recent advances in vision-language foundation models, such as CONCH~\cite{lu2024visual}, PLIP~\cite{huang2023visual}, MUSK~\cite{xiang2025vision}, Gecko~\cite{kapse2025gecko} and CPath-CLIP ~\cite{sun2025cpath}, have demonstrated significant efficacy in bridging visual morphology with clinical language for WSI analysis. Building upon these foundation models, Pathology VQA has emerged as a key task, with the most recent frameworks employing MLLMs to achieve complex reasoning. Initial efforts primarily address localized analysis at the patch or region level, as seen in models such as LLaVA-Med~\cite{li2023llavamed}, Quilt-LLaVA~\cite{seyfioglu2024quilt}, and PathChat~\cite{lu2024multimodal}. More recently, the focus has shifted toward slide-level diagnostics, where frameworks such as SlideChat~\cite{chen2025slidechat} and WSI-LLaVA~\cite{liang2025wsi} attempt to handle comprehensive queries by aggregating massive visual features from gigapixel images. To enhance reasoning logic, agent-based frameworks such as PathFinder~\cite{ghezloo2025pathfinder}, WSI-Agents~\cite{lyu2025inline} and CpathAgent~\cite{sun2025cpathagent} have been proposed to emulate a pathologist's workflow via iterative reasoning. While their dynamic navigation ensures structured evidence gathering, this sequential process may incur significant inference latency. Unlike exhaustive aggregation or iterative agents, we focus on reducing token redundancy in slide-level VQA by distilling a question-aligned subset of patches for efficient diagnostic reasoning.

\myparagraph{Information Bottleneck in Computational Pathology.} The Information Bottleneck (IB) principle \cite{tishby2000information} is an information-theoretic framework for learning, positing that an optimal model should learn a ``bottleneck'' representation that is maximally compressive of the input while retaining the maximum possible information about the downstream task \cite{alemi2017deep}. Due to its inherent ability to mitigate redundancy, the IB principle has been increasingly adopted in computational pathology to address domain-specific challenges \cite{li2023task, shi2023mg, gao2023childhood, zhang2024prototypical}. For example, \cite{li2023task} proposed a variational IB-based fine-tuning strategy to learn task-specific features for WSI classification. Concurrently, \cite{zhang2024prototypical} employed prototypical IB and information disentanglement to tackle the massive redundancy issues present in multimodal cancer survival prediction. Despite its demonstrated potential in classification and survival analysis, to the best of our knowledge, the IB framework has not yet been explored for pathology VQA. This represents a significant gap, as the hierarchical and token-intensive nature of LLM-based VQA models, which must process a massive number of visual tokens from WSIs, presents a critical challenge of information redundancy and computational inefficiency that the IB principle is ideally suited to address.

\section{Methods}
\label{sec:methods}

Our HistoSelect framework is shown in \Cref{fig:overview}. It mainly consists of two core components designed to select question-related patches for downstream multimodal reasoning.  The first part is tissue segmentation, where the WSI is partitioned into distinct spatial regions corresponding to $M$ different tissue types (e.g., tumor, stromal, lymphocyte, as illustrated) using pathologist-designed prompts. The second part is the hierarchical selector, which enhances model explainability and token efficiency by selecting patch tokens most relevant to the question. This hierarchical selector includes a group sampler and a patch selector, both taking the question embedding as input. The group sampler predicts the sampling rate for each tissue group, while the patch selector calculates a selection probability for every patch within the group. By ranking these probabilities, the selector extracts the top $K$ most relevant patch tokens. Subsequently, the selected patch tokens and the question tokens are passed to the LLM decoder for answer generation. We detail each component in the following sections.

\myparagraph{Preliminaries.}
Given a WSI and a question $Q$, the WSI is initially divided into a collection of $N$ non-overlapping patches. Each patch is encoded by a pretrained vision encoder to produce a set of features $\mathbf{X}=\{\mathbf{x}_1, \mathbf{x}_2, \dots, \mathbf{x}_N\}$, where $\mathbf{x}_i\in\mathbb{R}^d$ is the $d$-dimensional feature vector for the $i$-th patch. The question is encoded by a text encoder into a question feature $\mathbf{q}\in\mathbb{R}^d$. Furthermore, each patch $i$ is associated with a distinct tissue label $l_i$, obtained from the tissue segmentation stage. The collection of all tissue labels is denoted as $L=\{l_1, l_2, \dots, l_N\}$, where $l_i \in \{1, \dots, M\}$ represents one of the $M$ defined tissue types.

\subsection{Tissue Segmentation}
To identify the tissue regions relevant to the clinical question, we consult with expert pathologists. We design $M$ general prompts, denoted as $\mathcal{P} = \{p_1, p_2, \dots, p_M\}$, where each prompt $P_j$ is designed to generally represent a key histological component within the WSI.
Specifically, we utilize the visual encoder from CONCH \cite{lu2024visual} to obtain the patch features $\mathbf{X} = \{\mathbf{x}_1, \mathbf{x}_2, \dots, \mathbf{x}_N\}$, and its text encoder to embed the $M$ prompts into a feature space, yielding the tissue prompt features $\mathcal{T} = \{\mathbf{t}_1, \mathbf{t}_2, \dots, \mathbf{t}_M\}$, where $\mathbf{t}_j$ is the feature for prompt $P_j$. The tissue label $l_i$ for patch $i$ is determined by the highest cosine similarity between the patch feature $\mathbf{x}_i$ and all prompt features $\mathbf{t}_j$: 

\begin{equation}
l_i = \underset{j \in \{1, \dots, M\}}{\text{argmax}} \left( \frac{\mathbf{x}_i \cdot \mathbf{t}_j}{\|\mathbf{x}_i\| \cdot \|\mathbf{t}_j\|} \right)
\end{equation}

The resulting set of labels $L=\{l_1, l_2, \dots, l_N\}$ effectively partitions the WSI into $M$ distinct tissue regions, which form the basis for the hierarchical selection process. 

\subsection{Group Sampler}
\label{subsec:group_sampler}

The hierarchical selection process begins with the group sampler, which determines the importance of each tissue region relative to the input question. For each of the $M$ tissue groups, we first compute a group prototype feature $\mathbf{g}_j$ by applying average pooling over all patch features $x_i$ belonging to that group $j$. Let $\mathcal{I}_j$ be the set of indices for patches belonging to tissue group $j$. The group prototype $\mathbf{g}_j$ is defined as:
\begin{equation}
\mathbf{g}_j = \frac{1}{N_j} \sum_{i \in \mathcal{I}_j} \mathbf{x}_i
\end{equation}
where $N_j$ is the total number of patches in group $j$. The group sampler $\mathcal{F}_{\text{group}}$ then predicts a sampling rate $r_j$ for group $j$, indicating the importance of this group. This prediction is based on the concatenation of the group prototype $\mathbf{g}_j$ and the question feature $\mathbf{q}$, ensuring the sampling is context-aware. The sampling rate $r_j$ is constrained to $(0, 1)$ using the sigmoid function $\sigma(\cdot)$:
\begin{equation}
r_j = \sigma \left( \mathcal{F}_{\text{group}} ([\mathbf{g}_j; \mathbf{q}]) \right)
\end{equation}
where the group sampler $\mathcal{F}_{\text{group}}$ is implemented with two linear layers, and $r_j$ dictates the proportion of patches to be sampled from group $j$.

\subsection{Patch Selector}
\label{subsec:patch_selector}

The patch selector performs the final, fine-grained selection of relevant patches within each tissue group. Similar to the group sampler, the selection mechanism is driven by the question context. For every patch $i$, the patch selector $\mathcal{F}_{\text{patch}}$ predicts a selection probability $s_i$. This prediction is based on the concatenation of the individual patch feature $\mathbf{x}_i$ and the question feature $\mathbf{q}$:
\begin{equation}
s_i = \sigma \left( \mathcal{F}_{\text{patch}} ([\mathbf{x}_i; \mathbf{q}]) \right)
\end{equation}
where $s_i \in (0, 1)$ is the predicted probability that patch $i$ is relevant to question $Q$, and $\mathcal{F}_{\text{patch}}$ is implemented with two linear layers. For each tissue group $j$, the number of patches to be selected $k_j$ is determined by multiplying the group's predicted sampling rate $r_j$ by its total size $N_j$, followed by rounding up:
\begin{equation}
    k_j = \left\lceil r_j \cdot N_j \right\rceil
\end{equation}
Finally, we select the top $k_j$ features, denoted as $\mathbf{Z}_j$, from group $j$ by ranking all features $\mathbf{x}_i$ belonging to $\mathcal{I}_j$ based on their selection probability $s_i$. The complete set of selected patch features $\mathbf{Z}$ is the union of all selected features across the $M$ groups: $\mathbf{Z} = \bigcup_{j=1}^{M} \mathbf{Z}_j$.

\subsection{Hierarchical IB for Patch Selection}
\label{subsec:inforamtion_bottleneck}

Our learning objective is based on the IB theory \cite{tishby2000information}, which seeks to learn an optimal compressed representation \text{Z} of the input features \text{X} that retains maximal information about the ground truth answer \text{Y}. Given the question feature $\mathbf{q}$, this objective is formally written as:
\begin{equation}
    \mathcal{L}_{\text{IB}} = I(\text{Z}; \text{Y} \mid \mathbf{q}) - \beta I(\text{Z}; \text{X} \mid \mathbf{q})
\label{eq:ib_objective}
\end{equation}
where $I(\cdot; \cdot)$ denotes the mutual information, and $\beta$ is a Lagrangian multiplier balancing the trade-off between relevance and compression. To accommodate the hierarchical structure of WSIs, we model the selection process via a joint latent variable $\text{Z} = (\text{Z}_g, \text{Z}_p)$, where $\text{Z}_g$ and $\text{Z}_p$ denote the group-level and patch-level selection variables, respectively. By applying the chain rule, the complexity term $I(\text{Z}, \text{X} \mid \mathbf{q})$ is decomposed into a marginal group-level term and a conditional patch-level term:
\begin{equation}
    I(\text{Z}_g, \text{Z}_p; \text{X} \mid \mathbf{q}) = I(\text{Z}_g; \text{X} \mid \mathbf{q}) + I(\text{Z}_p; \text{X} \mid \text{Z}_g, \mathbf{q})
\label{eq:mi_decomposition}
\end{equation}
Since mutual information is computationally intractable, we adopt the Variational Information Bottleneck (VIB) framework \cite{tishby2000information} to derive a tractable bound. Inspired by recent hierarchical IB frameworks \cite{shi2023mg, gao2023childhood} designed for multi-scale WSIs, we introduce the hierarchical variational posteriors $p_{\phi_g}(\text{Z}_g \mid X, \mathbf{q})$ and $p_{\phi_p}(\text{Z}_p \mid X, \text{Z}_g, \mathbf{q})$ to model the group-level sampling and patch-level selection, respectively. Furthermore, we utilize the LLM as the variational decoder $p_{\theta}(Y \mid Z, \mathbf{q})$ to generate the final answer. By approximating the complexity terms in \Cref{eq:mi_decomposition} with their corresponding KL divergence upper bounds, the resulting hierarchical IB objective function is formulated as:
\begin{equation}
\begin{aligned}
\mathcal{J}_{\text{HIB}} = \mathbb{E}_{\mathcal{D}} \Bigl[ & \mathbb{E}_{Z \sim p_{\phi}} [\log p_{\theta}(Y \mid Z, \mathbf{q})] \\
& - \beta_g D_{\text{KL}}(p_{\phi_g}(Z_g \mid X, \mathbf{q}) \parallel p_g) \\
& - \beta_p D_{\text{KL}}(p_{\phi_p}(Z_p \mid X, Z_g, \mathbf{q}) \parallel p_p) \Bigr]
\end{aligned}
\label{eq:vib_objective}
\end{equation}
where $p_g$ and $p_p$ are the prior distributions. The hyperparameters $\beta_g$ and $\beta_p$ independently regulate the information flow at the group-level and patch-level granularities, respectively. The complete derivation of this hierarchical decomposition is provided in the supplementary material.

\subsection{Loss Function and Implementation}
\label{subsec:loss_function}

We define the loss function to be minimized as $\mathcal{L}_{\text{final}} = -\mathcal{J}_{\text{HIB}}$. Following the hierarchical decomposition in \Cref{eq:vib_objective}, the total loss is formulated as a weighted sum of the task-specific VQA loss and the hierarchical compression terms:
\begin{equation}
    \mathcal{L}_{\text{final}} = \mathcal{L}_{\text{VQA}} + \beta_g \mathcal{L}_{\text{group}} + \beta_p \mathcal{L}_{\text{patch}}
\end{equation}
where $\mathcal{L}_{\text{group}}$ and $\mathcal{L}_{\text{patch}}$ are the compression loss terms derived from the group sampler and patch selector, respectively. The $\beta_g$ and $\beta_p$ control the trade-off between task-relevant information and redundancy at each stage. 

\myparagraph{VQA Loss.} The VQA loss is the negative log-likelihood over the answer sequence:
$$\mathcal{L}_{\text{VQA}} = - \sum_{t=1}^{T} \log p_{\theta}(y_t \mid y_{<t}, \text{Z}, \mathbf{q})$$
where $\text{Y}=\{y_1, \dots, y_T\}$ is the ground truth answer sequence, and the total loss is averaged over the dataset $\mathcal{D}$.

\myparagraph{Group-Level Compression Loss.}
Weighted by $\beta_g$, this term regularizes the group sampler by minimizing the deviation of the predicted group sampling rate $r_j$ from a pseudo-prior parameter $p_j^g$. In this context, $r_j$ and $p_j^g$ are interpreted as the parameters of two Bernoulli distributions. The loss is averaged over the $M$ tissue groups:
\begin{equation}
\mathcal{L}_{\text{group}} = \frac{1}{M} \sum_{j=1}^{M} \mathbf{D}_{\text{KL}}(\mathbf{B}(r_j) \parallel \mathbf{B}(p_j^g))
\end{equation}
In practice, the pseudo-prior parameter $p_j^g$ is implemented as the cosine similarity between the group prototype $\mathbf{g}_j$ and the question feature $\mathbf{q}$.

\myparagraph{Patch-Level Compression Loss.}
Weighted by $\beta_p$, this term regularizes the patch selector by minimizing the deviation of the patch selection probability $s_i$ from a pseudo-prior parameter $p_i^p$. In this context, $s_i$ and $p_i^p$ are interpreted as the parameters of two Bernoulli distributions. The loss is averaged over the $N$ patches:
\begin{equation}
\mathcal{L}_{\text{patch}} = \frac{1}{N} \sum_{i=1}^{N} \mathbf{D}_{\text{KL}}(\mathbf{B}(s_i) \parallel \mathbf{B}(p_i^p))
\end{equation}
Similarly, the pseudo-prior parameter $p_i^p$ is the cosine similarity between the patch feature $\mathbf{x}_i$ and the question feature $\mathbf{q}$.
The KL divergence between two Bernoulli distributions with parameters $\pi$ and $p$ is defined as:
\begin{equation}
\mathbf{D}_{\text{KL}}(\mathbf{B}(\pi) \parallel \mathbf{B}(p)) = \pi \log \frac{\pi}{p} + (1-\pi) \log \frac{1-\pi}{1-p}
\end{equation}

\subsection{Differentiable Hard Selection}
\label{subsec:ste}

During training, the model must sample discrete patches to compute $\mathcal{L}_{\text{VQA}}$, yet this sampling operation is non-differentiable. To overcome this, we adopt the Straight-Through Estimator (STE) \cite{bengio2013estimating}. Specifically, a hard binary mask is applied during the forward pass to select the top-$k_j$ features for each group $j$. During the backward pass, gradients are propagated directly through the soft probabilities ($r_j$ and $s_i$), bypassing the discrete sampling step. This technique enables the entire pipeline, including both the group sampler and the patch selector, to be optimized end-to-end under the guidance of the VQA loss.

\section{Experiments and Results}
\label{sec:experiments_results}

\begin{table*}[ht]
\centering
\scriptsize
\setlength{\tabcolsep}{8pt}
\caption{Close-ended VQA accuracy (\%) across three benchmarks: SlideBench-VQA, WSI-Bench, and our In-house Ovarian dataset. 
Our method consistently achieves the highest accuracy across all task categories and obtains the best overall average performance.}
\begin{tabular}{llcccccccc}
\toprule

& & \multicolumn{3}{c}{SlideBench-VQA (TCGA)} & \multicolumn{3}{c}{WSI-Bench (Close)} & \multicolumn{1}{c}{In-house Ovarian} & \multirow{2}{*}{Average} \\

\cmidrule(lr){3-5} \cmidrule(lr){6-8} \cmidrule(lr){9-9}

Method & Input & Microscopy & Diagnosis & Clinical & Morphology & Diagnosis & Treatment & Diagnosis & \\
\midrule

GPT-4o & Thumbnail & 39.24 & 24.12 & 44.67 & 47.07 & 53.06 & 87.50 & — & 49.28 \\
Quilt-LLaVA\cite{seyfioglu2024quilt} & Thumbnail & 52.39 & 30.19 & 49.33 & 94.13 & 84.13 & 97.92 & 70.67 & 68.39 \\
LLaVA-Med \cite{li2023llavamed} & Thumbnail & 52.15 & 29.97 & 47.33 & 91.04 & 81.32 & 95.83 & 70.67 & 66.90 \\
SlideChat \cite{chen2025slidechat} & WSI & 83.15 & 71.36 & 75.33 & 91.34 & 82.15 & 93.75 & 69.33 & 80.88 \\
HistoSelect & WSI & \textbf{84.62} & \textbf{73.09} & \textbf{77.30} & \textbf{94.57} & \textbf{85.79} & \textbf{97.92} & \textbf{73.33} & \textbf{83.80} \\
\bottomrule
\end{tabular}
\label{tab:maintable1}
\end{table*}

\begin{table*}[t]
\centering
\scriptsize
\setlength{\tabcolsep}{8pt} 

\caption{Open-ended VQA performance on the WSI-Bench dataset. 
We evaluate two tasks: \textit{Report Generation} and \textit{domain-specific VQA}. 
Our method achieves the highest performance across all report-generation metrics and the best results on 5 of 6 domain-specific metrics.}
\begin{tabular}{lccccccccccc} 
\toprule

& \multicolumn{11}{c}{WSI-Bench} \\
\cmidrule(lr){2-12} 

& \multicolumn{5}{c}{Report Generation} & \multicolumn{2}{c}{Morphology} & \multicolumn{2}{c}{Diagnosis} & \multicolumn{2}{c}{Treatment} \\
\cmidrule(lr){2-6} \cmidrule(lr){7-8} \cmidrule(lr){9-10} \cmidrule(lr){11-12} 

\multicolumn{1}{c}{Method} & BLEU-1 & BLEU-2 & BLEU-3 & BLEU-4 & ROUGE-L & WSI-P & WSI-R & WSI-P & WSI-R & WSI-P & WSI-R \\
\midrule

GPT-4o & 0.202 & 0.069 & 0.030 & 0.016 & 0.132 & 0.220 & 0.204 & 0.472 & 0.457 & 0.513 & 0.704 \\
WSI-VQA \cite{chen2024wsivqa} & 0.301 & 0.225 & 0.181 & 0.155 & 0.343 & 0.395 & 0.462 & 0.436 & 0.525 & 0.591 & 0.595 \\
MI-Gen \cite{chen2024wsicaption} & 0.403 & 0.306 & 0.248 & 0.209 & 0.446 & — & — & — & — & — & — \\
Histo-Gen \cite{guo2024histgen} & 0.406 & 0.307 & 0.248 & 0.208 & 0.448 & — & — & — & — & — & — \\
Quilt-LLaVA \cite{seyfioglu2024quilt} & 0.421 & 0.316 & 0.257 & 0.216 & 0.455 & 0.453 & 0.484 & 0.521 & 0.552 & 0.751 & \textbf{0.807} \\
SlideChat \cite{chen2025slidechat} & 0.413 & 0.312 & 0.254 & 0.215 & 0.450 & 0.512 & 0.541 & 0.501 & 0.522 & 0.745 & 0.712 \\
HistoSelect & \textbf{0.431} & \textbf{0.324} & \textbf{0.262} & \textbf{0.221} & \textbf{0.463} & \textbf{0.538} & \textbf{0.589} & \textbf{0.542} & \textbf{0.587} & \textbf{0.766} & 0.801 \\
\bottomrule
\end{tabular}
\label{tab:maintable2} 
\end{table*}

\myparagraph{Datasets and Preprocessing.} We conduct experiments on three slide-level VQA datasets, including two public benchmarks and one private dataset. The public datasets include 1) \underline{SlideBench-VQA}~\cite{chen2025slidechat} comprises 4,560 WSIs and 176K VQA pairs, spanning 10 different cancer types and covering three different scenarios: Microscopy, Diagnosis, and Clinical.
2) \underline{WSI-Bench}~\cite{liang2025wsi} contains 9,850 WSIs and 180K VQA pairs, with scenarios focusing on Morphological analysis, Diagnosis, and Treatment Planning. Finally, to assess the model's generalizability and clinical robustness, we curate a 3) \underline{Private Ovarian Dataset}. This dataset consists of 375 WSIs and 375 corresponding VQA pairs annotated by pathologists, focusing on key diagnostic features of ovarian cancer, and is used as an independent test set. We followed the CLAM \cite{lu2021data} methodology for patch-cutting and feature extraction, processing all patches to a size of $224 \times 224$.

\myparagraph{Baselines.} We compare our method with several state-of-the-art WSI-based MLLMs. These baselines include models specifically designed for pathological and medical VQA tasks, such as Quilt-LLaVA \cite{seyfioglu2024quilt}, WSI-VQA \cite{chen2024wsivqa}, LLaVA-Med \cite{li2023llavamed}, and SlideChat \cite{chen2025slidechat}, as well as models specialized for WSI report generation, including MI-Gen \cite{chen2024wsicaption} and Hist-Gen \cite{guo2024histgen}. We also include the general-purpose MLLM GPT-4o as a non-specialized baseline. 

\myparagraph{Evaluation Metrics.}
For the closed-ended tasks, we group the questions based on various clinical categories. Performance is evaluated using accuracy.  
For the open-ended answer generation tasks, we adopt text-generation metrics, including BLEU and ROUGE-L to measure semantic similarity between generated and reference answers.  
Following WSI-LLaVA~\cite{liang2025wsi}, we additionally employ two \emph{LLM-as-a-judge} metrics: WSI-Precision (WSI-P), which evaluates the factual correctness of model responses, and WSI-Relevance (WSI-R), which assesses how well each response aligns with the reference answer in a clinical context.

\begin{figure*}[t]
\centering
\includegraphics[width=1.0\textwidth]{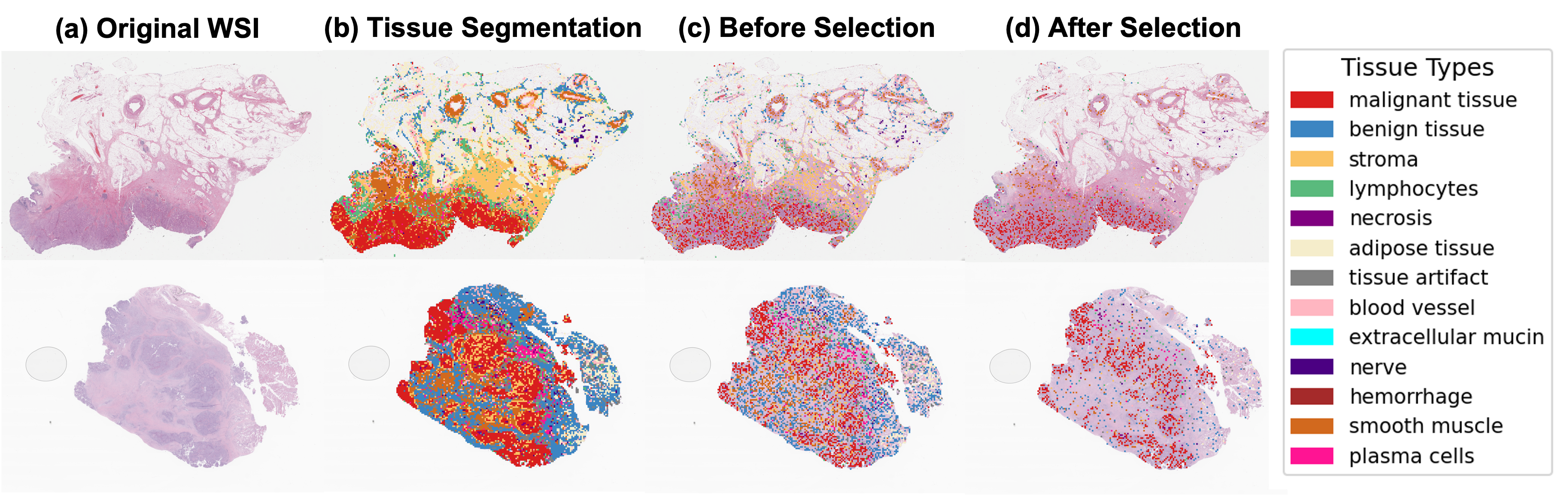}
\caption{Visualization of tissue segmentation and selection process. (a) Original WSI. (b) Tissue segmentation mask. (c)Visualization before selection (a randomly selected subset is shown for clarity). (d) Visualization after selection. Compared to (c), the patches selected by our model in (d) significantly remove non-tumor patches, demonstrating an improved focus on informative tumor-related regions.}
\label{fig:selection}
\end{figure*}

\myparagraph{Implementation Details.}
Following SlideChat \cite{chen2025slidechat}, we employ the CONCH encoder \cite{lu2024visual} to extract patch-level features and LongNet \cite{ding2023longnet, xu2024gigapath} for slide-level features, utilizing Qwen2.5-7B-Instruct \cite{team2024qwen2} as our LLM framework. We use the text encoder from CONCH to obtain question embeddings. For all experiments, we adhere to the official data splits of SlideBench-VQA \cite{chen2025slidechat} and WSI-Bench \cite{liang2025wsi}. Our training is conducted in two stages: consistent with SlideChat \cite{chen2025slidechat}, the first stage focuses on projector training for modality alignment. Subsequently, the second stage jointly fine-tunes the projector, the LLM (using LoRA), and our \ours module. Detailed hyperparameter settings are provided in the supplementary material.

\subsection{Quantitative Results}

\myparagraph{Close-ended Selection Performance.} Table~\ref{tab:maintable1} presents the close-ended VQA performance of our model against several state-of-the-art baselines across three benchmarks: SlideBench-VQA (TCGA), WSI-Bench (Close), and our In-house Ovarian dataset. We compare our WSI-based method against both thumbnail-based models (such as GPT-4o, Quilt-LLaVA) and other WSI-based models (such as SlideChat). The results clearly demonstrate that our model achieves the best performance across all tested categories, attaining an average score of 83.80\% and significantly outperforming all other baseline methods.

\myparagraph{Open-ended Generation Performance.} To evaluate the model's open-ended text generation capabilities, we conducted tests on the WSI-Bench benchmark, with results detailed in Table~\ref{tab:maintable2}. We use BLEU (1-4) and ROUGE-L metrics to assess the quality of Report Generation and WSI metrics to evaluate the domain-specific VQA performance. The experimental results demonstrate the significant advantages of our model. For Report Generation, our model achieves the highest scores across all five metrics, with its BLEU-4 (0.221) and ROUGE-L (0.463) scores notably surpassing other advanced models like Quilt-LLaVA and SlideChat. In the domain-specific VQA tasks, our model obtains the best results in 5 out of 6 metrics, proving its superior and well-balanced open-ended VQA capabilities.

\subsection{Qualitative Result}

\myparagraph{Visualization.} To intuitively demonstrate the effectiveness of our question-aware selection mechanism, we provide qualitative visualizations in Figure \ref{fig:selection}. The figure illustrates the model's workflow, proceeding from (a) WSI to (b) the tissue segmentation mask, (c) a subset of candidate patches extracted from the tissue regions, and (d) the sparse set of question-relevant patches after our model's selection. As shown, our method successfully filters out a large number of background and diagnostically irrelevant patches, allowing the model to focus its computation and attention on the most salient regions to answer the VQA query.

\myparagraph{Pathologist's Evaluation.} To validate the practical utility and interpretability of our model from a clinical perspective, we conducted a human evaluation survey with two independent pathologists. We mainly evaluate the model interpretability and performance from two aspects, with a detailed survey in the supplementary material:

\textbf{1. Tissue segmentation}, we presented the pathologists with the original slide and our generated tissue mask. They were asked to rate the following on a 5-point Likert scale (1 = Strongly Disagree, 5 = Strongly Agree):
\begin{itemize}
    \item Q1: ``How accurate is the tissue segmentation?"
\end{itemize}

\textbf{2. Patch selection}, we showed the pathologists the visualizations of patches before and after selection for a given question. They were asked to rate:
\begin{itemize}
    \item Q2.1: ``Does the model filter out a lot of question-irrelevant patches?"
    \item Q2.2: ``Are the selected patches sufficient to answer the question?"
\end{itemize}
The detailed average scores are presented in \Cref{tab:pathologist_scores_modified}. We are encouraged to find that the average rating for all four questions exceeded 3.5. This strongly indicates that (1) pathologists find our tissue segmentation accurate, and (2) they confirm that our selection model effectively filters out irrelevant regions while preserving the necessary diagnostic information to answer the clinical question.

\begin{table}[htbp]
\centering
\scriptsize

\caption{Average scores from the pathologist evaluation survey.}
\label{tab:pathologist_scores_modified} 
\setlength{\tabcolsep}{5pt} 
\begin{tabular}{lccc}
\toprule

Category & Question ID & Avg. Score (P1) & Avg. Score (P2) \\
\midrule

\multirow{1}{*}{Tissue Seg.} & Q1 Accuracy & 4.17 & 3.67\\
\midrule

\multirow{2}{*}{Patch Selection} & Q2.1 Question-relevant & 4.80 & 3.87 \\
& Q2.2 Answer-relevant & 4.67 & 3.73 \\
\bottomrule
\end{tabular}
\end{table}

\subsection{Ablation Studies}

To validate the effectiveness of \ours, we conduct a series of ablation studies. We analyze three key aspects of our model: (1) the superiority of our learned selection mechanism against alternative strategies, (2) the contribution of each component in our hierarchical framework, and (3) the impact of the token budget on model performance.

\begin{table*}[ht]
\parbox{.33\linewidth}{
\setlength{\tabcolsep}{2pt}
\centering
\scriptsize
\caption{Selection Mechanism Ablation.}
\begin{tabular}{lccc}
\toprule
Accuracy & Morphology & Diagnosis & Treatment \\
\midrule

Random Sampling & 88.84 & 78.02 & 91.67 \\
DivPrune & 90.01 & 80.99 & 93.75\\
Simple Similarity & 92.22 & 81.98 & 93.75 \\
Ours & \textbf{94.57} & \textbf{85.79} & \textbf{97.92} \\
 \bottomrule
\end{tabular}
\label{tab:ablation1}
}
\hfill
\parbox{.33\linewidth}{
\setlength{\tabcolsep}{2pt}
\centering
\scriptsize
\caption{Model Component Ablation Study.}
\begin{tabular}{lccc}
\toprule
Accuracy & Morphology & Diagnosis & Treatment \\
\midrule

Random Sampling  & 88.84 & 78.02 & 91.67 \\ 
w/o Group Sampler & 91.78 & 81.82 & 95.83\\ 
w/o Patch Selector & 92.07 & 81.32 & 93.75\\ 
Ours & \textbf{94.57} & \textbf{85.79} & \textbf{97.92}\\ 
 \bottomrule
\end{tabular}
\label{tab:ablation_components}
}
\hfill
\parbox{.33\linewidth}{
\setlength{\tabcolsep}{2pt}
\centering
\scriptsize
\caption{Ablation Study on $\#$ of Tokens.}
\begin{tabular}{lccc}
\toprule
Accuracy & Morphology & Diagnosis & Treatment \\
\midrule

10k Tokens & 94.12 & 85.12 & 97.92 \\
5k Tokens & \textbf{94.57} & \textbf{85.79} & \textbf{97.92} \\
2k Tokens & 93.83 & 83.80 & 95.83 \\
1k Tokens & 91.19 & 82.15 & 95.83 \\
 \bottomrule
\end{tabular}
\label{tab:ablation_tokens}
}
\end{table*}

\myparagraph{Selection Mechanism.} In Table~\ref{tab:ablation1}, we first compare our full model against alternative selection baselines, all constrained to the same token budget. The poor performance of random sampling serves as a lower bound, confirming that intelligent selection is critical. While the diversity-based method DivPrune \cite{alvar2025divprune} performs better, it still falls short of our model, indicating that selecting question-relevant patches is more important than simply selecting diverse ones. Most importantly, we evaluate a simple similarity baseline that replaces our learnable selectors ($\mathcal{F}_{\text{group}}$ and $\mathcal{F}_{\text{patch}}$) with the non-learnable pseudo-prior parameters $p_j^g$ and $p_i^p$ (derived from cosine similarity) directly. Its inferior performance strongly validates our core hypothesis: an end-to-end learned policy, regularized by the IB objective, is essential for identifying the most salient visual evidence and significantly outperforms static, similarity-based heuristics.

\myparagraph{Model Components.} In Table~\ref{tab:ablation_components}, we dissect our hierarchical architecture to understand each component. The baseline model, which defaults to Random Sampling, yields the poorest results. Removing only the patch selector ($\mathcal{F}_{\text{patch}}$) forces the model to rely on coarse group selection, and the subsequent performance drop highlights the necessity of fine-grained selection for critical patches. Conversely, removing the group sampler ($\mathcal{F}_{\text{group}}$) degrades the model to a ``flat'' selection mechanism, forcing the patch selector to search globally across all $N$ patches. Its poor performance relative to our full model confirms the benefit of our coarse-to-fine approach, as the group sampler effectively narrows the search space. The superior performance of our full model demonstrates that the group sampler and the patch selector work synergistically.

\paragraph{Impact of Token Budget.}
In Table~\ref{tab:ablation_tokens}, we analyze the performance of \ours\ under different token budget limits. We observe that performance improves as the token count limit increases from 1k to 5k, indicating the benefit of more visual context. However, the model achieves its peak performance at 5k tokens. Notably, increasing the limit further to 10k tokens provides no additional performance gain and even shows a slight degradation, likely due to the introduction of redundant information. This result is significant: \ours\ validates that a large portion of a WSI is redundant for a specific question, as it achieves its optimal accuracy by selecting a compact, sufficient subset of only 30\% of the total patches. This translates to a 70\% reduction in token computation while maximizing diagnostic accuracy. 

\section{Conclusion}
\label{conclusion}

In this work, we introduce HistoSelect, a question-aware framework addressing explainability and redundancy in current pathology VQA models. By mimicking the coarse-to-fine diagnostic strategy of human pathologists, HistoSelect efficiently prunes question-irrelevant patches, thereby increasing the signal-to-noise ratio for the downstream MLLM. This targeted selection process provides transparent and attributable visual evidence for prediction. Extensive experiments across diverse datasets, complemented by rigorous evaluations from practicing pathologists, confirm that HistoSelect not only achieves state-of-the-art performance but also delivers the trustworthy, explainable reasoning necessary to bridge the gap between automated analysis and clinical adoption in computational pathology.

\section*{Acknowledgment}

This work has been supported by Mayo Clinic Center for Individualized Medicine, and by the generous support of Schmidt Sciences and the Susan Morrow Legacy Foundation. This work was also partially supported by grants NSF CCF-2144901, NIH R01NS143143, and R01CA297843.

{
    \small
    \bibliographystyle{ieeenat_fullname}
    \bibliography{main}

@String(CVPR= {IEEE Conf. Comput. Vis. Pattern Recog.})

@String(ICCV= {Int. Conf. Comput. Vis.})

@String(ECCV= {Eur. Conf. Comput. Vis.})

@String(ICLR = {Int. Conf. Learn. Represent.})

@String(CVPR  = {CVPR})

@String(ICCV  = {ICCV})

@String(ECCV  = {ECCV})

@String(ICLR  = {ICLR})

@article{lu2021ai,
  title={AI-based pathology predicts origins for cancers of unknown primary},
  author={Lu, Ming Y and Chen, Tiffany Y and Williamson, Drew FK and Zhao, Melissa and Shady, Maha and Lipkova, Jana and Mahmood, Faisal},
  journal={Nature},
  year={2021}
}

@article{niazi2019digital,
  title={Digital pathology and artificial intelligence},
  author={Niazi, Muhammad Khalid Khan and Parwani, Anil V and Gurcan, Metin N},
  journal={The lancet oncology},
  year={2019}
}

@article{barisoni2020digital,
  title={Digital pathology and computational image analysis in nephropathology},
  author={Barisoni, Laura and Lafata, Kyle J and Hewitt, Stephen M and Madabhushi, Anant and Balis, Ulysses GJ},
  journal={Nature Reviews Nephrology},
  year={2020}
}

@inproceedings{ilse2018attention,
  title={Attention-based deep multiple instance learning},
  author={Ilse, Maximilian and Tomczak, Jakub and Welling, Max},
  booktitle={ICML},
  year={2018}
}

@inproceedings{li2021dual,
  title={Dual-stream multiple instance learning network for whole slide image classification with self-supervised contrastive learning},
  author={Li, Bin and Li, Yin and Eliceiri, Kevin W},
  booktitle={CVPR},
  year={2021}
}

@inproceedings{zhang2022dtfd,
  title={DTFD-MIL: Double-tier feature distillation multiple instance learning for histopathology whole slide image classification},
  author={Zhang, Hongrun and Meng, Yanda and Zhao, Yitian and Qiao, Yihong and Yang, Xiaoyun and Coupland, Sarah E and Zheng, Yalin},
  booktitle={CVPR},
  year={2022}
}

@inproceedings{shao2021transmil,
  title={Transmil: Transformer based correlated multiple instance learning for whole slide image classification},
  author={Shao, Zhuchen and Bian, Hao and Chen, Yang and Wang, Yifeng and Zhang, Jian and Ji, Xiangyang and others},
  booktitle={NeurIPS},
  year={2021}
}

@inproceedings{qu2023boosting,
  title={Boosting whole slide image classification from the perspectives of distribution, correlation and magnification},
  author={Qu, Linhao and Yang, Zhiwei and Duan, Minghong and Ma, Yingfan and Wang, Shuo and Wang, Manning and Song, Zhijian},
  booktitle={CVPR},
  year={2023}
}

@inproceedings{tang2023multiple,
  title={Multiple instance learning framework with masked hard instance mining for whole slide image classification},
  author={Tang, Wenhao and Huang, Sheng and Zhang, Xiaoxian and Zhou, Fengtao and Zhang, Yi and Liu, Bo},
  booktitle={CVPR},
  year={2023}
}

@inproceedings{zhang2023attention,
  title={Attention-challenging multiple instance learning for whole slide image classification},
  author={Zhang, Yunlong and Li, Honglin and Sun, Yuxuan and Zheng, Sunyi and Zhu, Chenglu and Yang, Lin},
  booktitle={ECCV},
  year={2024}
}

@inproceedings{chen2021multimodal,
  title={Multimodal co-attention transformer for survival prediction in gigapixel whole slide images},
  author={Chen, Richard J and Lu, Ming Y and Weng, Wei-Hung and Chen, Tiffany Y and Williamson, Drew FK and Manz, Trevor and Shady, Maha and Mahmood, Faisal},
  booktitle={ICCV},
  year={2021}
}

@inproceedings{xu2023multimodal,
  title={Multimodal optimal transport-based co-attention transformer with global structure consistency for survival prediction},
  author={Xu, Yingxue and Chen, Hao},
  booktitle={ICCV},
  year={2023}
}

@inproceedings{zhou2023cross,
  title={Cross-modal translation and alignment for survival analysis},
  author={Zhou, Fengtao and Chen, Hao},
  booktitle={ICCV},
  year={2023}
}

@inproceedings{song2024multimodal,
  title={Multimodal Prototyping for cancer survival prediction},
  author={Song, Andrew H and Chen, Richard J and Jaume, Guillaume and Vaidya, Anurag J and Baras, Alexander S and Mahmood, Faisal},
  booktitle={ICML},
  year={2024}
}

@inproceedings{zhang2024prototypical,
  title={Prototypical Information Bottlenecking and Disentangling for Multimodal Cancer Survival Prediction},
  author={Zhang, Yilan and Xu, Yingxue and Chen, Jianqi and Xie, Fengying and Chen, Hao},
  booktitle={ICLR},
  year={2024}
}

@inproceedings{qu2022bi,
  title={Bi-directional weakly supervised knowledge distillation for whole slide image classification},
  author={Qu, Linhao and Wang, Manning and Song, Zhijian and others},
  booktitle={NeurIPS},
  year={2022}
}

@inproceedings{lin2023interventional,
  title={Interventional bag multi-instance learning on whole-slide pathological images},
  author={Lin, Tiancheng and Yu, Zhimiao and Hu, Hongyu and Xu, Yi and Chen, Chang-Wen},
  booktitle={CVPR},
  year={2023}
}

@inproceedings{zhu2025dgr,
  title={DGR-MIL: Exploring Diverse Global Representation in Multiple Instance Learning for Whole Slide Image Classification},
  author={Zhu, Wenhui and Chen, Xiwen and Qiu, Peijie and Sotiras, Aristeidis and Razi, Abolfazl and Wang, Yalin},
  booktitle={ECCV},
  year={2024}
}

@inproceedings{seyfioglu2024quilt,
  title={Quilt-llava: Visual instruction tuning by extracting localized narratives from open-source histopathology videos},
  author={Seyfioglu, Mehmet Saygin and Ikezogwo, Wisdom O and Ghezloo, Fatemeh and Krishna, Ranjay and Shapiro, Linda},
  booktitle={CVPR},
  year={2024}
}

@inproceedings{chen2025slidechat,
  title={Slidechat: A large vision-language assistant for whole-slide pathology image understanding},
  author={Chen, Ying and Wang, Guoan and Ji, Yuanfeng and Li, Yanjun and Ye, Jin and Li, Tianbin and Hu, Ming and Yu, Rongshan and Qiao, Yu and He, Junjun},
  booktitle={CVPR},
  year={2025}
}

@inproceedings{liang2025wsi,
  title={WSI-LLAVA: a multimodal large language model for whole slide image},
  author={Liang, Yuci and Lyu, Xinheng and Chen, Wenting and Ding, Meidan and Zhang, Jipeng and He, Xiangjian and Wu, Song and Xing, Xiaohan and Yang, Sen and Wang, Xiyue and others},
  booktitle={ICCV},
  year={2025}
}

@inproceedings{alvar2025divprune,
  title={Divprune: Diversity-based visual token pruning for large multimodal models},
  author={Alvar, Saeed Ranjbar and Singh, Gursimran and Akbari, Mohammad and Zhang, Yong},
  booktitle={CVPR},
  year={2025}
}

@article{lu2021data,
  title={Data-efficient and weakly supervised computational pathology on whole-slide images},
  author={Lu, Ming Y and Williamson, Drew FK and Chen, Tiffany Y and Chen, Richard J and Barbieri, Matteo and Mahmood, Faisal},
  journal={Nature Biomedical Engineering},
  year={2021}
}

@inproceedings{chen2024wsivqa,
  title={Wsi-vqa: Interpreting whole slide images by generative visual question answering},
  author={Chen, Pingyi and Zhu, Chenglu and Zheng, Sunyi and Li, Honglin and Yang, Lin},
  booktitle={ECCV},
  year={2024}
}

@inproceedings{chen2024wsicaption,
  title={Wsicaption: Multiple instance generation of pathology reports for gigapixel whole-slide images},
  author={Chen, Pingyi and Li, Honglin and Zhu, Chenglu and Zheng, Sunyi and Shui, Zhongyi and Yang, Lin},
  booktitle={MICCAI},
  year={2024}
}

@inproceedings{li2023llavamed,
  title={Llava-med: Training a large language-and-vision assistant for biomedicine in one day},
  author={Li, Chunyuan and Wong, Cliff and Zhang, Sheng and Usuyama, Naoto and Liu, Haotian and Yang, Jianwei and Naumann, Tristan and Poon, Hoifung and Gao, Jianfeng},
  booktitle={NeurIPS},
  year={2023}
}

@inproceedings{guo2024histgen,
  title={Histgen: Histopathology report generation via local-global feature encoding and cross-modal context interaction},
  author={Guo, Zhengrui and Ma, Jiabo and Xu, Yingxue and Wang, Yihui and Wang, Liansheng and Chen, Hao},
  booktitle={MICCAI},
  year={2024}
}

@article{lu2024visual,
  title={A visual-language foundation model for computational pathology},
  author={Lu, Ming Y and Chen, Bowen and Williamson, Drew FK and Chen, Richard J and Liang, Ivy and Ding, Tong and Jaume, Guillaume and Odintsov, Igor and Le, Long Phi and Gerber, Georg and others},
  journal={Nature medicine},
  year={2024}
}

@inproceedings{ding2023longnet,
  title={LongNet: Scaling Transformers to 1,000,000,000 Tokens},
  author={Ding, Jiayu and Ma, Shuming and Dong, Li and Zhang, Xingxing and Huang, Shaohan and Wang, Wenhui and Wei, Furu},
  booktitle={ICLR},
  year={2023}
}

@article{xu2024gigapath,
  title={A whole-slide foundation model for digital pathology from real-world data},
  author={Xu, Hanwen and Usuyama, Naoto and Bagga, Jaspreet and Zhang, Sheng and Rao, Rajesh and Naumann, Tristan and Wong, Cliff and Gero, Zelalem and González, Javier and Gu, Yu and Xu, Yanbo and Wei, Mu and Wang, Wenhui and Ma, Shuming and Wei, Furu and Yang, Jianwei and Li, Chunyuan and Gao, Jianfeng and Rosemon, Jaylen and Bower, Tucker and Lee, Soohee and Weerasinghe, Roshanthi and Wright, Bill J. and Robicsek, Ari and Piening, Brian and Bifulco, Carlo and Wang, Sheng and Poon, Hoifung},
  journal={Nature},
  year={2024}
}

@article{team2024qwen2,
  title={Qwen2 technical report},
  author={Team, Qwen and others},
  journal={arXiv preprint arXiv:2407.10671},
  year={2024}
}

@article{tishby2000information,
  title={The information bottleneck method},
  author={Tishby, Naftali and Pereira, Fernando C and Bialek, William},
  journal={arXiv preprint physics/0004057},
  year={2000}
}

@inproceedings{alemi2017deep,
  title={Deep Variational Information Bottleneck},
  author={Alemi, Alexander A and Fischer, Ian and Dillon, Joshua V and Murphy, Kevin},
  booktitle={ICLR},
  year={2017}
}

@inproceedings{li2023task,
  title={Task-specific fine-tuning via variational information bottleneck for weakly-supervised pathology whole slide image classification},
  author={Li, Honglin and Zhu, Chenglu and Zhang, Yunlong and Sun, Yuxuan and Shui, Zhongyi and Kuang, Wenwei and Zheng, Sunyi and Yang, Lin},
  booktitle={CVPR},
  year={2023}
}

@inproceedings{ikezogwo2023quilt,
  title={Quilt-1m: One million image-text pairs for histopathology},
  author={Ikezogwo, Wisdom and Seyfioglu, Saygin and Ghezloo, Fatemeh and Geva, Dylan and Sheikh Mohammed, Fatwir and Anand, Pavan Kumar and Krishna, Ranjay and Shapiro, Linda},
  booktitle={NeurIPS},
  year={2023}
}

@inproceedings{sun2025cpathagent,
  title={CPathAgent: An Agent-based Foundation Model for Interpretable High-Resolution Pathology Image Analysis Mimicking Pathologists' Diagnostic Logic},
  author={Sun, Yuxuan and Si, Yixuan and Zhu, Chenglu and Zhang, Kai and Shui, Zhongyi and Ding, Bowen and Lin, Tao and Yang, Lin},
  booktitle={NeurIPS},
  year={2025}
}

@article{sun2025pathbench,
  title={PathBench: Advancing the Benchmark of Large Multimodal Models for Pathology Image Understanding at Patch and Whole Slide Level},
  author={Sun, Yuxuan and Wu, Hao and Zhu, Chenglu and Si, Yixuan and Chen, Qizi and Zhang, Yunlong and Zhang, Kai and Li, Jingxiong and Cai, Jiatong and Wang, Yuhan and others},
  journal={TMI},
  year={2025}
}

@article{bengio2013estimating,
  title={Estimating or propagating gradients through stochastic neurons for conditional computation},
  author={Bengio, Yoshua and L{\'e}onard, Nicholas and Courville, Aaron},
  journal={arXiv preprint arXiv:1308.3432},
  year={2013}
}

@article{xu2021predicting,
  title={Predicting axillary lymph node metastasis in early breast cancer using deep learning on primary tumor biopsy slides},
  author={Xu, Feng and Zhu, Chuang and Tang, Wenqi and Wang, Ying and Zhang, Yu and Li, Jie and Jiang, Hongchuan and Shi, Zhongyue and Liu, Jun and Jin, Mulan},
  journal={Frontiers in oncology},
  year={2021}
}

@article{huang2023visual,
  title={A visual--language foundation model for pathology image analysis using medical twitter},
  author={Huang, Zhi and Bianchi, Federico and Yuksekgonul, Mert and Montine, Thomas J and Zou, James},
  journal={Nature medicine},
  year={2023}
}

@article{xiang2025vision,
  title={A vision--language foundation model for precision oncology},
  author={Xiang, Jinxi and Wang, Xiyue and Zhang, Xiaoming and Xi, Yinghua and Eweje, Feyisope and Chen, Yijiang and Li, Yuchen and Bergstrom, Colin and Gopaulchan, Matthew and Kim, Ted and others},
  journal={Nature},
  year={2025}
}

@inproceedings{sun2025cpath,
  title={Cpath-omni: A unified multimodal foundation model for patch and whole slide image analysis in computational pathology},
  author={Sun, Yuxuan and Si, Yixuan and Zhu, Chenglu and Gong, Xuan and Zhang, Kai and Chen, Pingyi and Zhang, Ye and Shui, Zhongyi and Lin, Tao and Yang, Lin},
  booktitle={CVPR},
  year={2025}
}

@article{lu2024multimodal,
  title={A multimodal generative AI copilot for human pathology},
  author={Lu, Ming Y and Chen, Bowen and Williamson, Drew FK and Chen, Richard J and Zhao, Melissa and Chow, Aaron K and Ikemura, Kenji and Kim, Ahrong and Pouli, Dimitra and Patel, Ankush and others},
  journal={Nature},
  year={2024}
}

@inproceedings{ghezloo2025pathfinder,
  title={Pathfinder: A multi-modal multi-agent system for medical diagnostic decision-making applied to histopathology},
  author={Ghezloo, Fatemeh and Seyfioglu, Mehmet Saygin and Soraki, Rustin and Ikezogwo, Wisdom O and Li, Beibin and Vivekanandan, Tejoram and Elmore, Joann G and Krishna, Ranjay and Shapiro, Linda},
  booktitle={CVPR},
  year={2025}
}

@inproceedings{lyu2025inline,
  title={WSI-Agents: A Collaborative Multi-agent System for Multi-modal Whole Slide Image Analysis},
  author={Lyu, Xinheng and Liang, Yuci and Chen, Wenting and Ding, Meidan and Yang, Jiaqi and Huang, Guolin and Zhang, Daokun and He, Xiangjian and Shen, Linlin},
  booktitle={MICCAI},
  year={2025}
}

@article{shi2023mg,
  title={MG-trans: Multi-scale graph transformer with information bottleneck for whole slide image classification},
  author={Shi, Jiangbo and Tang, Lufei and Gao, Zeyu and Li, Yang and Wang, Chunbao and Gong, Tieliang and Li, Chen and Fu, Huazhu},
  journal={TMI},
  year={2023}
}

@article{gao2023childhood,
  title={Childhood leukemia classification via information bottleneck enhanced hierarchical multi-instance learning},
  author={Gao, Zeyu and Mao, Anyu and Wu, Kefei and Li, Yang and Zhao, Liebin and Zhang, Xianli and Wu, Jialun and Yu, Lisha and Xing, Chao and Gong, Tieliang and others},
  journal={TMI},
  year={2023}
}

@inproceedings{huang2024hard,
  title={Hard negative sample mining for whole slide image classification},
  author={Huang, Wentao and Hu, Xiaoling and Abousamra, Shahira and Prasanna, Prateek and Chen, Chao},
  booktitle={MICCAI},
  year={2024}
}

@inproceedings{kapse2025gecko,
  title={Gecko: Gigapixel vision-concept contrastive pretraining in histopathology},
  author={Kapse, Saarthak and Pati, Pushpak and Yellapragada, Srikar and Das, Srijan and Gupta, Rajarsi R and Saltz, Joel and Samaras, Dimitris and Prasanna, Prateek},
  booktitle={ICCV},
  year={2025}
}

@inproceedings{zhang20252dmamba,
  title={2dmamba: Efficient state space model for image representation with applications on giga-pixel whole slide image classification},
  author={Zhang, Jingwei and Nguyen, Anh Tien and Han, Xi and Trinh, Vincent Quoc-Huy and Qin, Hong and Samaras, Dimitris and Hosseini, Mahdi S},
  booktitle={CVPR},
  year={2025}
}

@inproceedings{xu2025match,
  title={MATCH: Multi-faceted Adaptive Topo-Consistency for Semi-Supervised Histopathology Segmentation},
  author={Xu, Meilong and Hu, Xiaoling and Abousamra, Shahira and Li, Chen and Chen, Chao},
  booktitle={NeurIPS},
  year={2025}
}

@inproceedings{xu2024semi,
  title={Semi-supervised segmentation of histopathology images with noise-aware topological consistency},
  author={Xu, Meilong and Hu, Xiaoling and Gupta, Saumya and Abousamra, Shahira and Chen, Chao},
  booktitle={ECCV},
  year={2024}
}

@misc{streamlit,
  title = {Streamlit},
  year = {2025},
  howpublished = {\url{https://streamlit.io/}}
}
}

\newpage

\twocolumn[
\centering
\Large
\textbf{Act Like a Pathologist: Tissue-Aware Whole Slide Image Reasoning} \\\vspace{0.05cm}{--- Supplementary Material ---}
\\
\vspace{1.5em}
]

\noindent In this supplementary material, we provide additional technical derivations, experimental details, and qualitative analyses to complement the main manuscript. First, \Cref{sec:supp1} presents the mathematical derivation of our Hierarchical Information Bottleneck objective. Subsequently, we introduce the details of our in-house ovarian dataset and the specialized evaluation tool developed for pathologist assessment in \Cref{sec:supp2} and \Cref{sec:supp3}, respectively. We then elaborate on the implementation details of our two-stage training strategy in \Cref{sec:supp4}. To further demonstrate the model's performance, \Cref{sec:supp5} provides additional quantitative results on public benchmarks, followed by extensive qualitative visualizations on both public and private datasets in \Cref{sec:supp6}. A comprehensive ablation study investigating various training configurations and sampling distributions is presented in \Cref{sec:supp7}. Finally, we discuss the limitations of our current approach and suggest future research directions in \Cref{sec:supp8}.

\section{Hierarchical IB Objective Derivation}
\label{sec:supp1}

\subsection{Variational Information Bound}
By definition, the mutual information $I(A; B)$ between two random variables $A$ and $B$ is:
\begin{equation}
I(A; B) = \mathbb{E}_{p(A, B)} \left[ \log \frac{p(A \mid B)}{p(A)} \right]
\end{equation}
In practice, the true marginal distribution $p(A)$ is typically intractable. To derive a computable bound, we introduce a variational prior $q(A)$. Utilizing the non-negativity of the KL divergence:
\begin{equation}
D_{\text{KL}}(p(A) \parallel q(A)) = \mathbb{E}_{p(A)} \left[ \log \frac{p(A)}{q(A)} \right] \geq 0
\end{equation}
it follows that:
\begin{equation}
\mathbb{E}_{p(A)}[\log p(A)] \geq \mathbb{E}_{p(A)}[\log q(A)]
\end{equation}
Consequently, by substituting this inequality into the definition of $I(A; B)$, we arrive at the variational upper bound for the compression term, following the VIB framework \cite{alemi2017deep}:
\begin{equation}
I(A; B) \leq \mathbb{E}_{p(B)} [D_{\text{KL}}(p(A \mid B) \parallel q(A))]
\label{eq:supp_vib_lemma}
\end{equation}

\subsection{Hierarchical Variational Decomposition}
As stated in the main text, the total compression term $I(Z; X \mid \mathbf{q})$ is decomposed into group-level and patch-level terms using the chain rule for mutual information:
\begin{equation}
I(Z_g, Z_p; X \mid \mathbf{q}) = I(Z_g; X \mid \mathbf{q}) + I(Z_p; X \mid Z_g, \mathbf{q})
\end{equation}
By applying the variational bound from \Cref{eq:supp_vib_lemma} to each hierarchical component, we derive the tractable hierarchical constraints for our selection process. For the group-level complexity, we introduce the variational posterior $p_{\phi_g}(Z_g \mid X, \mathbf{q})$ and the group-level prior $p_g(Z_g \mid \mathbf{q})$. The mutual information term is then bounded by:
\begin{equation}
I(Z_g; X \mid \mathbf{q}) \leq \mathbb{E}_{\mathcal{D}} [D_{\text{KL}}(p_{\phi_g}(Z_g \mid X, \mathbf{q}) \parallel p_g)]
\end{equation}
Similarly, for the patch-level complexity, we introduce the conditional posterior $p_{\phi_p}(Z_p \mid X, Z_g, \mathbf{q})$ and its corresponding conditional prior $p_p(Z_p \mid Z_g, \mathbf{q})$. The mutual information term is then bounded by:
\begin{equation}
\begin{aligned}
I(Z_p; X \mid Z_g, \mathbf{q}) \leq & \mathbb{E}_{\mathcal{D}} \Bigl[ \mathbb{E}_{Z_g \sim p_{\phi_g}} [D_{\text{KL}}(p_{\phi_p}(Z_p \mid X, Z_g, \mathbf{q}) \\
& \parallel p_p(Z_p \mid Z_g, \mathbf{q}))] \Bigr]
\end{aligned}
\end{equation}

\subsection{Deriving the Final Objective}
For the relevance term $I(Z; Y \mid \mathbf{q})$, which measures the predictive power of the selected features for the answer $Y$, we utilize the LLM as a variational decoder $p_{\theta}(Y \mid Z, \mathbf{q})$ to obtain a tractable lower bound:
\begin{equation}
I(Z; Y \mid \mathbf{q}) \geq \mathbb{E}_{\mathcal{D}} \mathbb{E}_{Z \sim p_{\phi}} [\log p_{\theta}(Y \mid Z, \mathbf{q})]
\end{equation}
By substituting the complexity upper bounds and the relevance lower bound into the original IB objective $\mathcal{L}_{\text{IB}}$, we arrive at a tractable training objective. In the standard VIB \cite{alemi2017deep} formulation, a single hyperparameter $\beta$ typically regulates the entire compression term. To better accommodate the hierarchical nature of WSIs and provide greater flexibility in hyperparameter tuning, we extend this formulation by assigning independent Lagrange multipliers, $\beta_g$ and $\beta_p$, as group-level and patch-level regularizers, respectively. This decoupling allows the model to independently regulate the information bottleneck at each granularity. The resulting HIB objective is formulated as:
\begin{equation}
\begin{aligned}
\mathcal{J}_{\text{HIB}} &= \mathbb{E}_{\mathcal{D}} \Bigl[ \mathbb{E}_{Z \sim p_{\phi}} [\log p_{\theta}(Y \mid Z, \mathbf{q})] \\
& \quad - \beta_g D_{\text{KL}}(p_{\phi_g}(Z_g \mid X, \mathbf{q}) \parallel p_g) \\
& \quad - \beta_p \mathbb{E}_{Z_g \sim p_{\phi_g}} [D_{\text{KL}}(p_{\phi_p}(Z_p \mid X, Z_g, \mathbf{q}) \parallel p_p)] \Bigr]
\end{aligned}
\label{eq:supp_hib_objective}
\end{equation}
In practice, the nested expectation over $Z_g$ in the patch-level term is empirically estimated through the group-level sampling process. By using the specific sampling rate made by the group sampler during the forward pass, the theoretical objective reduces to the empirical loss function presented in Equation (8) of the main text.

\begin{figure*}[t]
    \centering
    \includegraphics[width=0.95\textwidth]{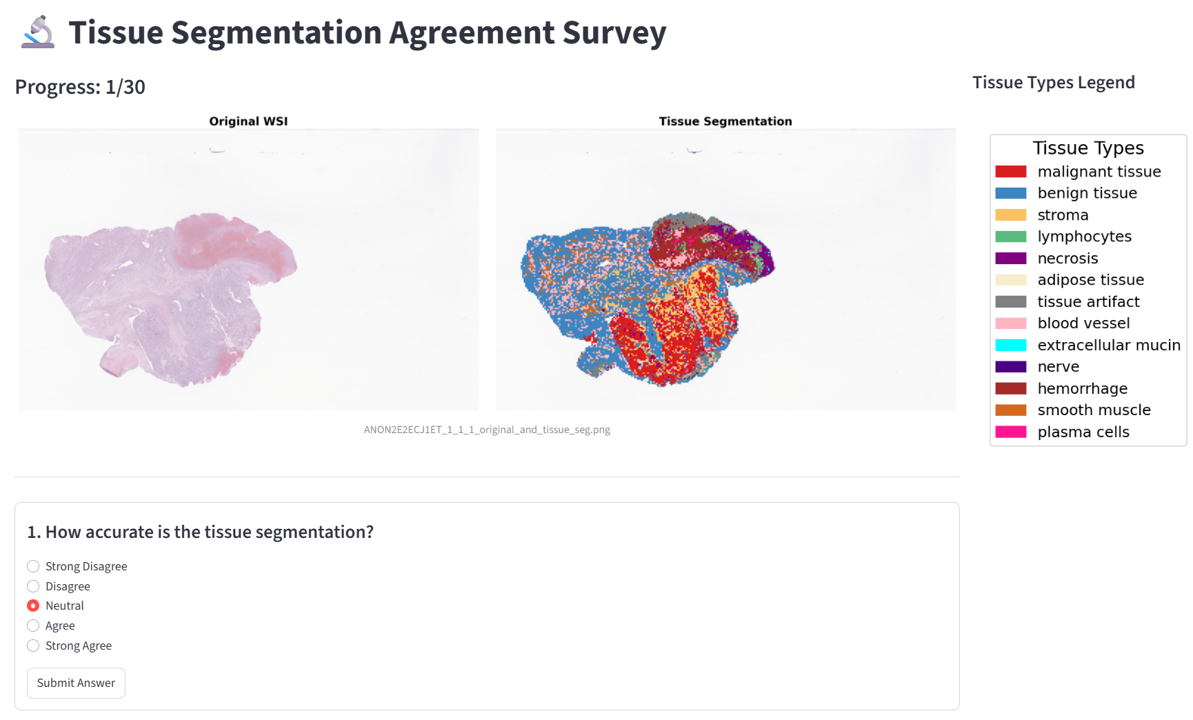}
    \caption{The user interface for the Tissue Segmentation Survey. The central area shows a side-by-side comparison of the original WSI and the tissue segmentation result. The right legend clarifies the tissue classes, and the bottom section collects the pathologists' rating.}
    \label{fig:survey_tissue}
\end{figure*}

\section{In-house Ovarian Dataset}
\label{sec:supp2}
To demonstrate the generalizability of our proposed model, we curated a small-scale, in-house ovarian dataset. This dataset is compiled from WSIs of ovarian tissues and formatted into question-answer pairs, focusing on distinct histological phenotypes visible within the WSIs. The dataset includes four primary diagnostic categories, based on the observed tumor morphology. In total, the dataset comprises 375 question-answer pairs. The distribution of samples across the four categories is as follows: endometrioid ($n=81$), clear cell carcinoma ($n=82$), high grade serous carcinoma ($n=123$), and serous borderline carcinoma ($n=89$).

\begin{figure*}[t]
    \centering
    \includegraphics[width=0.9\textwidth]{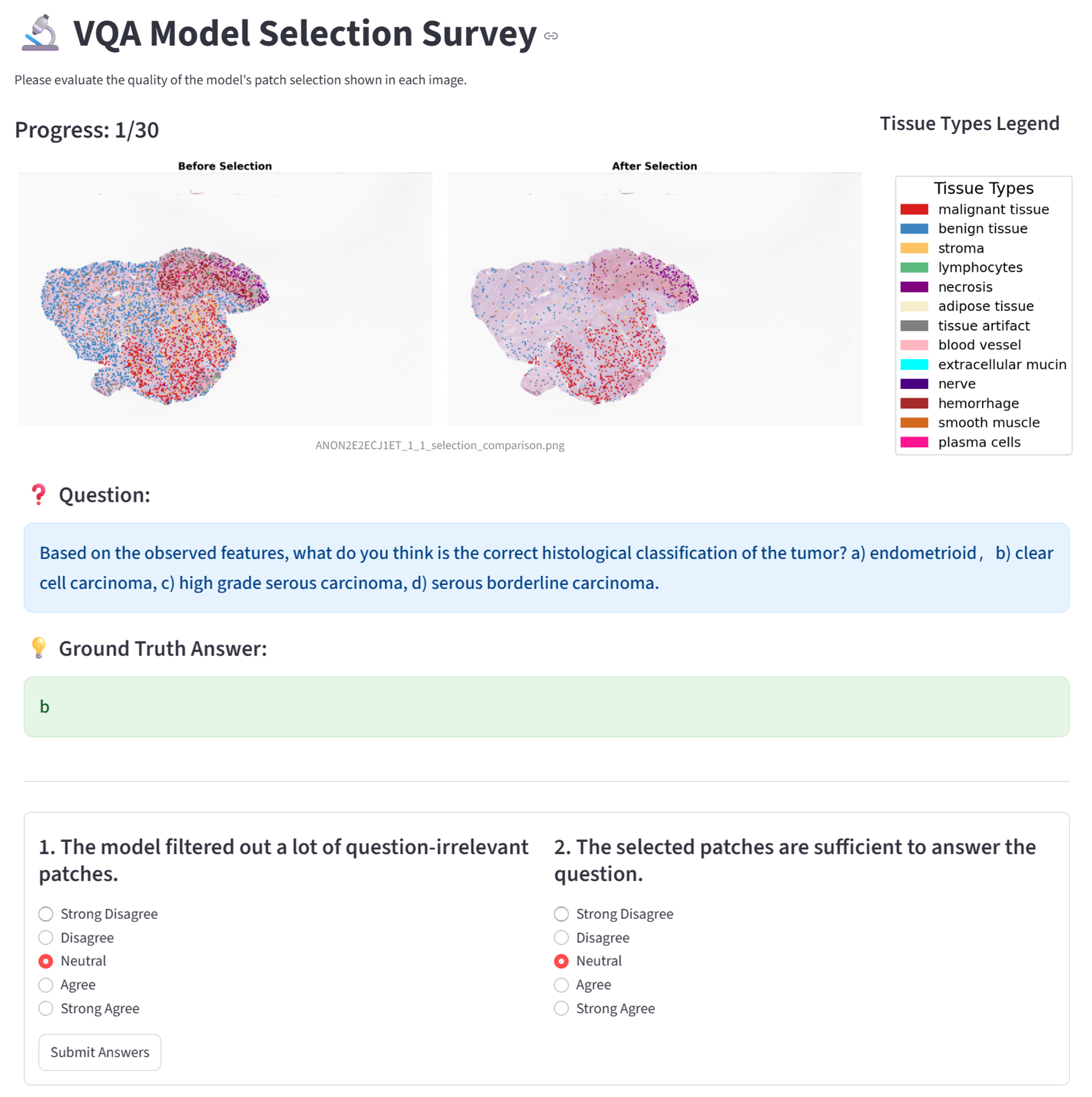}
    \caption{The user interface for the Patch Selection Survey. The view visualizes the ``Before'' and ``After'' states of patch selection. Note that the input question and ground truth answer are displayed below the images to provide necessary context for the pathologists' evaluation.}
    \label{fig:survey_vqa}
\end{figure*}

A typical question within the dataset is structured as a multiple-choice classification task based on visual features observed in the WSI. An example is provided below:
\begin{quote}
    \textbf{Example Question-Answer Pair:} Based on the observed features, what do you think is the correct histological classification of the tumor?
    \begin{enumerate}[(a)]
        \item endometrioid
        \item clear cell carcinoma
        \item high grade serous carcinoma
        \item serous borderline carcinoma
    \end{enumerate}
\end{quote}

\section{Evaluation Tool}
\label{sec:supp3}

To verify the reliability of our tissue segmentation and ensure that the model selects question-relevant patches for inference, we developed an interactive survey tool based on Streamlit~\cite{streamlit}. This evaluation is twofold: it primarily validates the accuracy of tissue segmentation, followed by an assessment of the patch selection performance. We detail these two components in the following sections.

\myparagraph{Tissue Segmentation Verification.} In the first part, we focus on the tissue segmentation results. As illustrated in \Cref{fig:survey_tissue}, the user interface features a side-by-side comparison view: the original WSI is displayed on the left, while the corresponding model-generated segmentation mask is shown on the right. To assist pathologists in interpreting the results, a color-coded legend is provided on the sidebar, mapping specific colors to tissue types (e.g., red for malignant tissue, yellow for stroma). 

Pathologists are asked to verify the alignment between the WSI and the mask using these visual cues. Based on their inspection, they rate the segmentation quality via a radio button selection:
\begin{itemize}
    \item \textbf{Q1:} ``How accurate is the tissue segmentation?'' (1 = Strongly Disagree, 5 = Strongly Agree)
\end{itemize} After the user clicks ``Submit Answer'', the tool automatically logs the feedback and advances to the next sample.

\myparagraph{Patch Selection Assessment.}
In the second part, the tool adapts to evaluate the model's coarse-to-fine selection capability. As shown in \Cref{fig:survey_vqa}, we visualize the patches in both their ``Before'' and ``After'' selection states. 

To ensure sufficient diagnostic context for evaluation, the specific ``Question'' (highlighted in blue) and the ``Ground Truth Answer'' (highlighted in green) are explicitly displayed below the images. Pathologists assess whether the model successfully removes irrelevant regions while retaining diagnostic information by answering the following:
\begin{itemize}
    \item \textbf{Q2.1:} ``Does the model filter out a significant number of question-irrelevant patches?''
    \item \textbf{Q2.2:} ``Are the selected patches sufficient to answer the question?''
\end{itemize} To conduct a robust evaluation, we randomly sampled a total of 30 cases, comprising 20 from the public dataset and 10 from the private dataset. Two independent pathologists were invited to perform the annotations using the developed tool. To ensure the reliability and objectivity of the assessment, we calculated the average scores from both annotators as the final metric for pathologist-based evaluation.

\begin{table*}[t]
\centering
\scriptsize
\setlength{\tabcolsep}{4pt} 
\caption{Quantitative results on SlideBench-VQA (TCGA). Performance is evaluated using accuracy and macro-averaged metrics.}
\begin{tabular}{lcccc|cccc|cccc}
\toprule
& \multicolumn{4}{c}{\textbf{Microscopy}} & \multicolumn{4}{c}{\textbf{Diagnosis}} & \multicolumn{4}{c}{\textbf{Clinical}} \\
\cmidrule(lr){2-5} \cmidrule(lr){6-9} \cmidrule(lr){10-13}
Method & Acc. & Macro-P & Macro-R & Macro-F1 & Acc. & Macro-P & Macro-R & Macro-F1 & Acc. & Macro-P & Macro-R & Macro-F1 \\
\midrule
GPT-4o & 39.24 & 36.12 & 39.45 & 35.54 & 24.12 & 22.83 & 24.34 & 21.72 & 44.67 & 42.93 & 44.95 & 43.58 \\
Quilt-LLaVA \cite{seyfioglu2024quilt} & 52.39 & 46.75 & 50.09 & 46.84 & 30.19 & 27.94 & 30.34 & 27.15 & 49.33 & 47.01 & 48.91 & 47.54 \\
LLaVA-Med \cite{li2023llavamed} & 52.15 & 46.30 & 49.80 & 46.51 & 29.97 & 27.84 & 29.93 & 26.86 & 47.33 & 45.12 & 47.19 & 45.58 \\
SlideChat \cite{chen2025slidechat} & 83.15 & \textbf{81.77} & 78.50 & 79.70 & 71.36 & 71.80 & 65.22 & 67.52 & 75.33 & 73.84 & 72.74 & 72.98 \\
\midrule
\textbf{Ours} & \textbf{84.62} & 80.79 & \textbf{80.47} & \textbf{80.33} & \textbf{73.09} & \textbf{72.10} & \textbf{68.08} & \textbf{69.22} & \textbf{77.30} & \textbf{73.97} & \textbf{74.24} & \textbf{73.94} \\
\bottomrule
\end{tabular}
\label{tab:slidebench_results}
\end{table*}

\begin{table*}[t]
\centering
\scriptsize
\setlength{\tabcolsep}{4pt}
\caption{Quantitative results on WSI-Bench (Close-ended). Performance is evaluated using accuracy and macro-averaged metrics.}
\begin{tabular}{lcccc|cccc|cccc}
\toprule
& \multicolumn{4}{c}{\textbf{Morphology}} & \multicolumn{4}{c}{\textbf{Diagnosis}} & \multicolumn{4}{c}{\textbf{Treatment (Binary)}} \\
\cmidrule(lr){2-5} \cmidrule(lr){6-9} \cmidrule(lr){10-13}
Method & Acc. & Macro-P & Macro-R & Macro-F1 & Acc. & Macro-P & Macro-R & Macro-F1 & Acc. & Macro-P & Macro-R & Macro-F1 \\
\midrule
GPT-4o & 47.07 & 42.89 & 47.29 & 43.19 & 53.06 & 49.02 & 53.37 & 49.27 & 87.50 & 79.12 & 83.76 & 81.05 \\
Quilt-LLaVA \cite{seyfioglu2024quilt} & 94.13 & 91.74 & 91.19 & 91.42 & 84.13 & 81.35 & 78.68 & 79.63 & \textbf{97.92} & \textbf{98.75} & 94.44 & 96.42 \\
LLaVA-Med \cite{li2023llavamed} & 91.04 & 86.21 & 87.59 & 86.83 & 81.32 & 77.80 & 74.81 & 76.02 & 95.83 & 93.16 & 93.16 & 93.16 \\
SlideChat \cite{chen2025slidechat} & 91.34 & 86.11 & 87.92 & 86.97 & 82.15 & 79.15 & 75.58 & 77.02 & 93.75 & 88.68 & 91.88 & 90.16 \\
\midrule
\textbf{Ours} & \textbf{94.57} & \textbf{92.66} & \textbf{91.34} & \textbf{91.85} & \textbf{85.79} & \textbf{85.03} & \textbf{79.45} & \textbf{81.70} & \textbf{97.92} & 95.00 & \textbf{98.72} & \textbf{96.72} \\
\bottomrule
\end{tabular}
\label{tab:wsibench_results}
\end{table*}

\section{Implementation Details}
\label{sec:supp4}
In the first stage, following Slidechat \cite{chen2025slidechat}, the projector and slide encoder are set to be trainable while the remaining components are frozen. This stage utilizes the WSI-caption data for initial alignment, employing a learning rate of 1e-3 for 3 epochs. In the second stage, we train the entire model for 2 epochs with a learning rate of 1e-4 and a batch size of 1. Specifically, we apply LoRA to the LLM to ensure efficient parameter updates. For the hyperparameters for our group sampler and patch selector, we employ a linear warmup schedule for the first 5000 iterations. During this warmup, the $\beta_g$ weight increases from 0 to 0.1, and the $\beta_p$ weight increases from 0 to 0.2, then they are held constant.

\begin{figure}[htbp]
\centering
\includegraphics[width=\columnwidth]{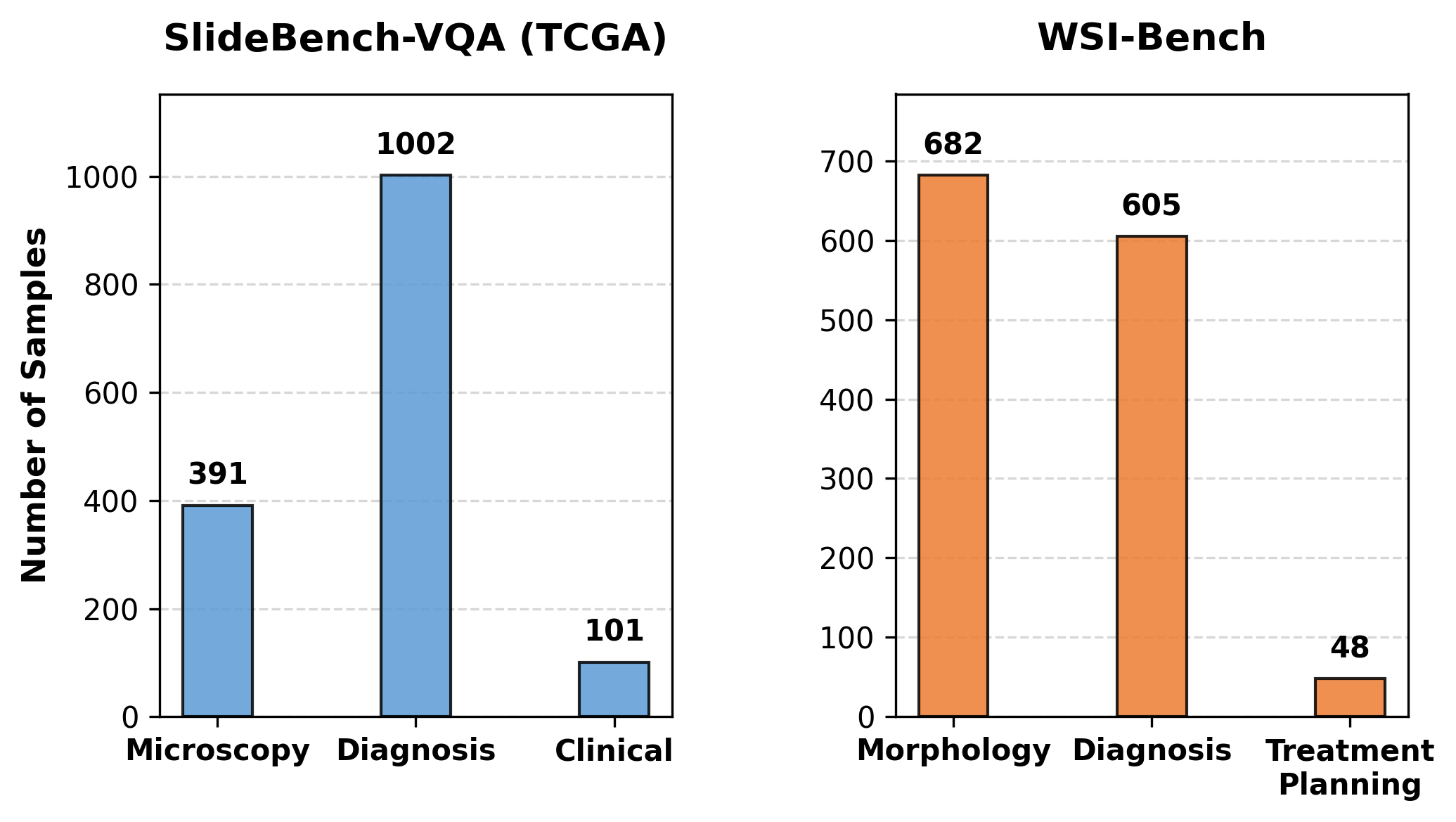}
\caption{The sample distribution for the test set.}
\label{fig:distribution}
\end{figure}

\section{Additional Quantitative Results.} 
\label{sec:supp5}
Figure~\ref{fig:distribution} illustrates the sample distribution across different task categories in SlideBench-VQA and WSI-Bench. To further validate the reliability of \ours, we report detailed macro-averaged metrics (Macro-Precision, Macro-Recall, and Macro-F1) across these two major benchmarks. As shown in Table~\ref{tab:slidebench_results} and Table~\ref{tab:wsibench_results}, our method consistently achieves superior performance in these balanced metrics on both SlideBench-VQA and WSI-Bench. These improvements demonstrate that our approach effectively identifies task-relevant patches and generalizes well across various categories. 

\begin{figure*}[t]
\centering
\includegraphics[width=1.0\textwidth]{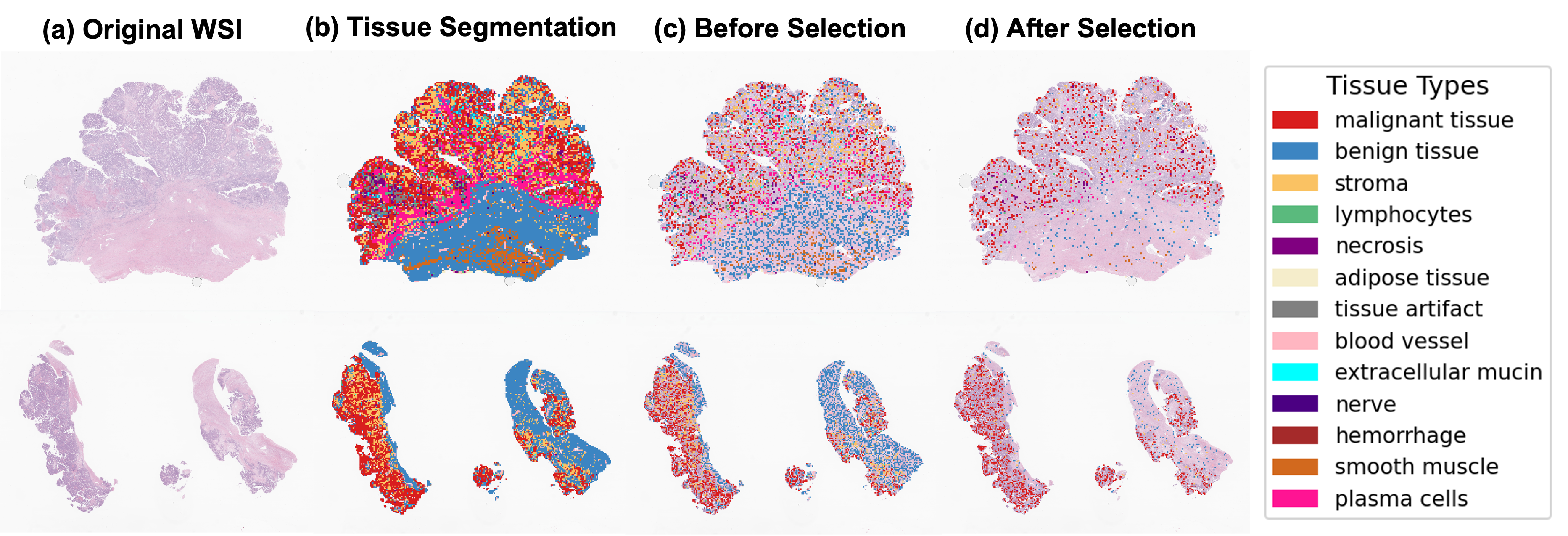}
\caption{Additional visualization of the selection process on the public WSI-Bench dataset. (a) Original WSI. (b) The tissue segmentation mask. (c) A visualization of candidate patches extracted from tissue regions prior to selection. (d) The sparse set of patches retained by our model. As observed in (d), the model effectively suppresses irrelevant regions, focusing the attention solely on the informative patches required for the VQA task.}
\label{fig:supp_public}
\end{figure*}

\begin{figure*}[t]
\centering
\includegraphics[width=1.0\textwidth]{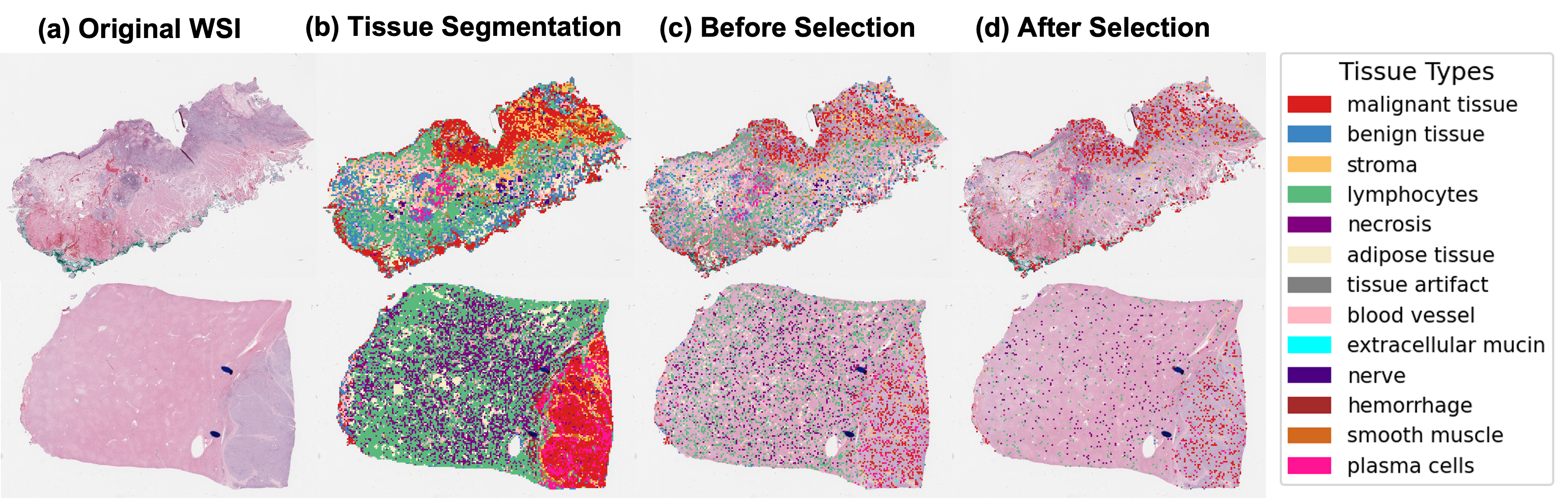}
\caption{Visualization of the selection process on the private Ovarian dataset. The figure follows the same pipeline as the main manuscript and Figure~\ref{fig:supp_public}: (a) Original WSI. (b) Tissue segmentation mask. (c) Patches before selection. (d) Patches after selection. These results demonstrate the robustness of our method against domain shifts common in private clinical data. The model successfully filters out non-informative tissue, preserving only the regions essential for accurate question answering.}
\label{fig:supp_private}
\end{figure*}

\section{Additional Qualitative Results}
\label{sec:supp6}

To provide a more comprehensive evaluation of our proposed method, we present additional qualitative visualizations in this section. Due to the page constraints of the main manuscript, we extend our analysis here to demonstrate the model's performance across different data distributions. Specifically, we visualize the effectiveness of our question-aware selection mechanism on both the public TCGA dataset and an in-house Ovarian dataset. These results further validate that our method can consistently filter out background noise and diagnostically irrelevant patches, enabling the model to focus efficiently on the regions most related to the VQA query.

\subsection{Results on Public Dataset}
Figure \ref{fig:supp_public} showcases the visualization results on the public TCGA dataset from WSI-Bench \cite{liang2025wsi}. Consistent with the findings in the main text, our model demonstrates strong generalization capabilities. It successfully identifies and retains key histological features required to answer the question while discarding a significant portion of irrelevant and redundant patches, thereby ensuring that the downstream reasoning is primarily driven by the most informative patches.

\subsection{Results on Private Dataset}
To assess the robustness of our model in a real-world clinical setting, Figure \ref{fig:supp_private} illustrates the selection process on our private ovarian dataset. Despite potential domain shifts such as variations in staining protocols and scanner properties compared to the public dataset, our question-aware selector maintains high precision. It effectively selects informative regions relevant to the query, verifying the method's applicability to proprietary clinical workflows.

\subsection{Sampling Rate Distribution Analysis}
\label{sec:sampling_dist}
\begin{figure}[htbp]
\centering
\includegraphics[width=0.88\columnwidth]{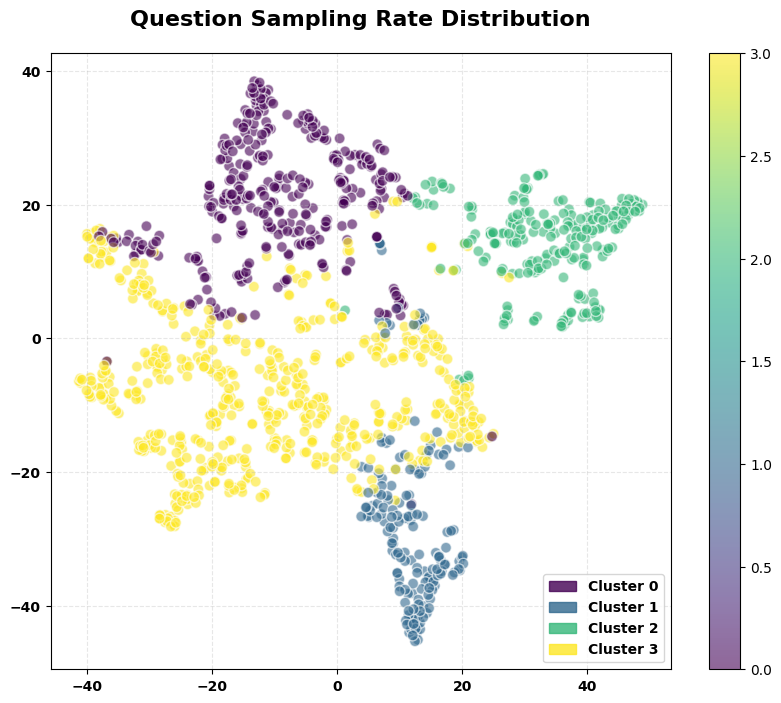}
\caption{Visualization of question-aware sampling distributions. We visualize the 13-dimensional sampling rate vectors from the WSI-Bench Diagnosis and Morphology test set using t-SNE.}
\label{fig:tsne_clusters}
\end{figure}

To investigate the sampling rate distribution across different questions, we conducted a quantitative analysis on the Diagnosis and Morphology subset of the WSI-Bench dataset \cite{liang2025wsi}. Specifically, we represented the tissue group sampling rate for each question as a 13-dimensional vector, where each dimension corresponds to the normalized sampling rate of a specific tissue component. We then applied K-means clustering to these vectors and visualized the resulting groupings using t-SNE, as shown in Figure \ref{fig:tsne_clusters}. The emergence of four distinct clusters ($K=4$) demonstrates that our model generates diverse and structured sampling patterns in the feature space. This grouping behavior confirms that the selection mechanism effectively navigates the complex composition of WSIs by adaptively prioritizing different histological tissue types based on the semantic focus of the question, rather than collapsing into a fixed, question-agnostic distribution.

\begin{figure*}[htbp]
\centering
\includegraphics[width=0.9\textwidth]{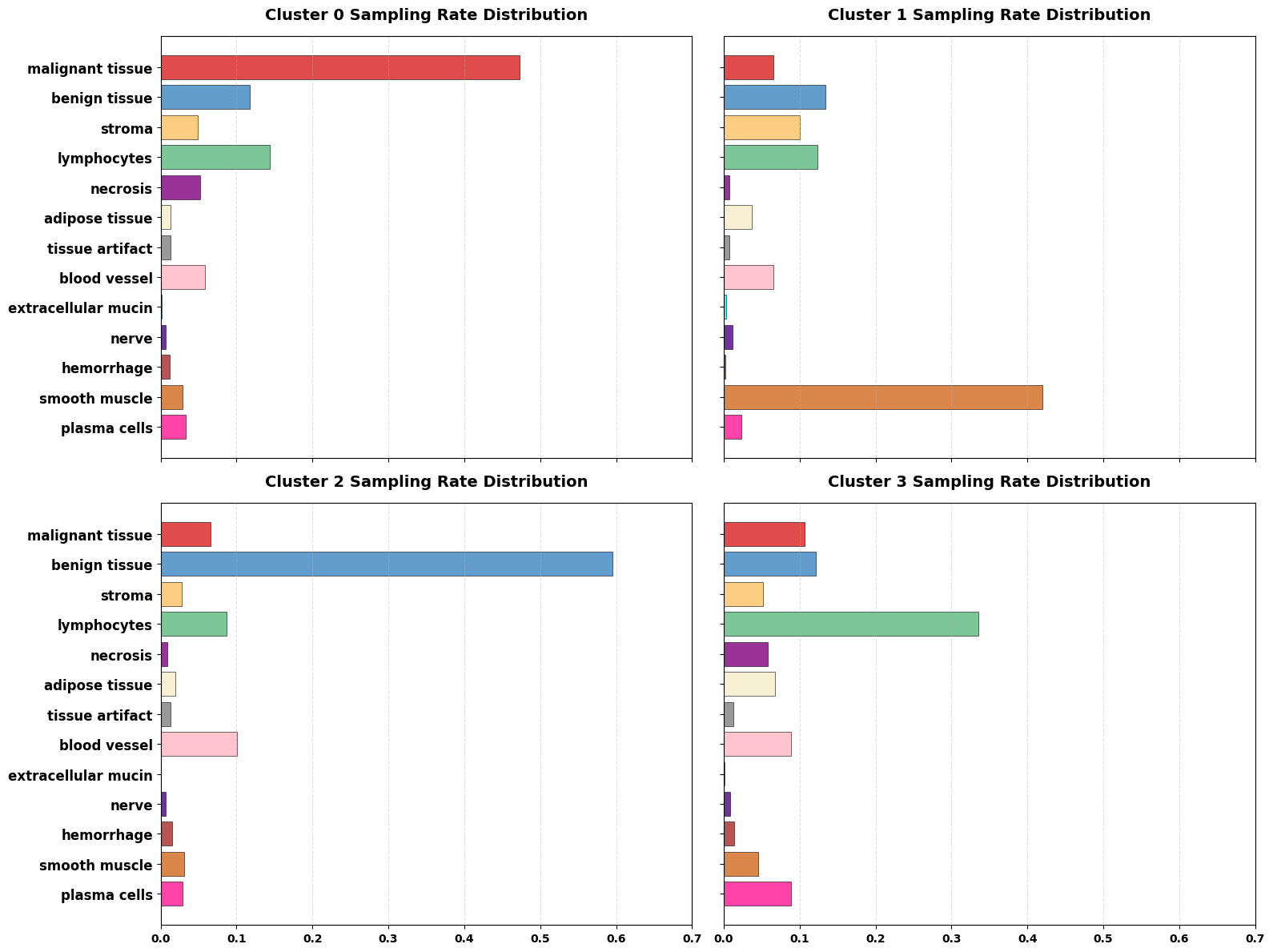}
\caption{Mean sampling rate distribution bar charts for identified clusters. We report the average 13-dimensional sampling vectors for the four clusters discovered in Fig. \ref{fig:tsne_clusters}. Each cluster exhibits a unique sampling rate pattern.}
\label{fig:cluster_barcharts}
\end{figure*}

The cluster-specific sampling distributions, as visualized in Fig.~\ref{fig:cluster_barcharts}, illustrate that our model develops distinct, task-driven sampling patterns. Each cluster corresponds to a specific clinical focus in the questions:

\begin{itemize}
    \item \textbf{Cluster 0 (Tumor Classification)}: Prioritizes malignant tissue, aligning with questions focused on histological tumor grading and classification. \\
    \textit{Example: Based on the observed features, what do you think is the correct histological classification of the tumor? A) Adenocarcinoma B) Small cell carcinoma C) Squamous cell carcinoma D) Large cell carcinoma}

    \item \textbf{Cluster 1 (Cellular Morphology)}: Shifts focus toward smooth muscle and stromal components, providing the necessary context for evaluating cellular variability and mitotic activity. \\
    \textit{Example: What are the notable features of the cellular morphology in this slide? A) Nuclei are uniform in appearance, showing no signs of active division. B) There is minimal variability in nuclear size, with a low rate of cell division. C) Nuclei appear extremely pleomorphic, with a very high rate of mitotic activity. D) There is moderate variability in nuclear size and shape, with a moderate rate of cell division and presence of single cells.}

    \item \textbf{Cluster 2 (Tissue Architecture)}: Shows a strong preference for benign tissue and surrounding structures, which is essential for assessing the overall microanatomy and glandular patterns. \\
    \textit{Example: What observations can you make about the tissue architecture on this slide? A) The tumor forms well-organized acinar structures with a clear glandular pattern. B) The tumor is characterized by prominent chicken-wire vasculature providing stroma. C) Tumor cells create extensive solid sheets, with a completely homogeneous pattern. D) The tissue maintains normal microanatomy with minimal deviation.}

    \item \textbf{Cluster 3 (Tumor Infiltration)}: Concentrates on lymphocytes and extracellular components, capturing the critical interface where tumor cells infiltrate the stroma and adipose layers. \\
    \textit{Example: What is the observed pattern of tumor infiltration in this specimen? A) Tumor cells are limited to the submucosal layer without muscularis propria involvement. B) Tumor cells infiltrate the stroma, extending into the muscularis propria and adipose tissue. C) Tumor cells remain within glandular structures without stromal invasion. D) There is only infiltration into the adipose tissue, sparing the submucosal layer.}
\end{itemize}
This clear divergence confirms that our selection mechanism is question-aware. Instead of relying on a static saliency map, the model dynamically re-prioritizes different histological tissue types based on the semantic intent of the question, ensuring that the most relevant patches are selected for each specific question.

\section{Ablation Study}
\label{sec:supp7}

In this section, we conduct extensive ablation studies to evaluate the effectiveness of the proposed components in \ours. We first analyze the impact of hyperparameters $\beta_g$ and $\beta_p$ in our loss function, which control the information bottleneck at the \textit{group sampler} and \textit{patch selector} levels, respectively. Subsequently, we investigate the influence of different training strategies and assess the model-agnostic generalization of our selector modules across various base models. Finally, a group selection analysis is performed to demonstrate the critical role of group-level selection in handling complex multi-tissue reasoning tasks.

\begin{table}[htbp]
\centering
\scriptsize
\begin{tabular}{lccc} 
\toprule
$\beta_g$ & Morphology & Diagnosis & Treatment \\
\midrule
0 & 91.78 & 81.82 & 95.83 \\
0.1 & 93.39 & 84.13 & 95.83 \\
0.2 & \textbf{94.57} & \textbf{85.79} & \textbf{97.92} \\
0.3 & 93.10 & 83.25 & 93.75 \\
\bottomrule
\end{tabular}
\caption{Ablation Study on the weight $\beta_g$ for the group sampler.}
\label{tab:ablation_beta_g}
\end{table}

\myparagraph{Impact of $\beta_g$ for the Group Sampler.}
Table \ref{tab:ablation_beta_g} reports the model performance under different values of $\beta_g \in \{0, 0.1, 0.2, 0.3\}$ while keeping $\beta_p$ fixed. We observe that when $\beta_g = 0$, the group sampler lacks the necessary regularization to filter out irrelevant tissue groups, leading to a lower signal-to-noise ratio and suboptimal performance. As $\beta_g$ increases to $0.2$, the model effectively suppresses background noise at the group level, achieving the best overall accuracy across all tasks. However, further increasing $\beta_g$ to $0.3$ results in a performance drop. This suggests that an overly aggressive penalty causes the sampler to excessively reduce the sampling rate of tissue groups that contain necessary contextual information.

\begin{table}[htbp]
\centering
\scriptsize
\begin{tabular}{lccc} 
\toprule
$\beta_p$ & Morphology & Diagnosis & Treatment \\
\midrule
0 & 92.07 & 81.32 & 93.75 \\
0.05 & 93.25 & 84.23 & 95.83 \\
0.10 & \textbf{94.57} & \textbf{85.79} & \textbf{97.92} \\
0.15 & 93.83 & 83.74 & 95.83 \\
\bottomrule
\end{tabular}
\caption{Ablation study on the weight $\beta_p$ for the patch selector.}
\label{tab:ablation_beta_p}
\end{table}

\myparagraph{Impact of $\beta_p$ for the Patch Selector.}
Table \ref{tab:ablation_beta_p} examines the effect of the patch-level weight $\beta_p \in \{0, 0.05, 0.10, 0.15\}$. Similar to the group level, setting $\beta_p = 0$ tends to retain a large number of redundant patches, which introduces potential interference for the answer predictor. We find that setting $\beta_p = 0.10$ yields the optimal balance, allowing the model to identify the most distinct and discriminative patches without losing critical information. Conversely, setting $\beta_p$ too high (e.g., $0.15$) leads to over-pruning, where the model is penalized for retaining informative patches, causing a loss of fine-grained details essential for accurate diagnosis and treatment prediction.

\begin{table}[htbp]
\centering
\scriptsize
\setlength{\tabcolsep}{4pt} 
\begin{tabular}{lcccc} 
\toprule
Training Strategy & Morphology & Diagnosis & Treatment & Avg. \\
\midrule
Joint Training  & 94.57 & 85.79 & 97.92 & 92.76 \\
Joint + Patch Selector & 94.71 & 85.95 & 97.92 & 92.86 \\
Joint + Group Sampler & \textbf{95.15} & \textbf{86.11} & \textbf{97.92} & \textbf{93.06} \\
\bottomrule
\end{tabular}
\caption{Ablation on training strategies.}
\label{tab:training_strategy}
\end{table}

\myparagraph{Impact of Training Strategies.} 
Beyond hyperparameter tuning, we investigate whether extending the training procedure further impacts performance. As shown in Table \ref{tab:training_strategy}, while the initial joint training yields strong results, conducting an additional epoch of training specifically for the sampler or selector modules proves beneficial. In particular, performing one extra epoch for the group sampler achieves the highest performance across most tasks. This suggests that after the joint training stage has established a solid foundation, the group sampler can further benefit from a dedicated optimization phase.

\myparagraph{Ablation with Different Base Models.} To verify the model-agnostic effectiveness of \ours, we conduct an experiment using Gemini 3 Flash as a frozen reasoning engine. Specifically, we randomly sample 200 cases from WSI-Bench and compare two input strategies: (1) a baseline using 100 randomly sampled patches with the question, and (2) using the top-100 patches selected by \ours with the same question. As shown in Table~\ref{tab:ablation_gemini}, our method yields a consistent performance boost even for a stronger foundation model. This demonstrates our model's ability to filter redundant noise and identify task-relevant tokens independently of the base model's reasoning capacity.

\begin{table}[htbp]
\centering
\scriptsize
\setlength{\tabcolsep}{6pt} 
\begin{tabular}{lcccc} 
\toprule
Method & ACC & Macro-P & Macro-R & Macro-F1 \\
\midrule
Gemini 3 Flash & 59.5 & 57.0 & 60.0 & 57.0 \\
Gemini 3 Flash + 
\ours & \textbf{62.5} & \textbf{58.0} & \textbf{64.0} & \textbf{59.0} \\
\bottomrule
\end{tabular}
\caption{Performance comparison on WSI-Bench (200 samples).}
\label{tab:ablation_gemini} 
\end{table}

\myparagraph{Impact of Group Selection.} 
We further conduct an analysis by comparing our base model with a version using ``Ideal Group Selection'' (i.e., perfect identification of relevant tissue regions). As reported in Table~\ref{tab:ablation_group}, better group selection consistently leads to higher performance. Notably, the gain is more significant for multi-tissue questions (e.g., tumor infiltration patterns) compared to single-tissue ones (e.g., tumor detection). This highlights that the group sampler is particularly essential for handling complex clinical reasoning that requires cross-tissue contextual integration.

\begin{table}[htbp]
\centering
\scriptsize
\setlength{\tabcolsep}{10pt}
\begin{tabular}{lccc} 
\toprule
Question Type & Base & Base + Ideal Group & Gain ($\Delta$) \\
\midrule
Single-tissue (Easy) & 85.0 & 87.0 & +2.0 \\
Multi-tissue (Hard)  & 73.0 & 79.0 & +6.0 \\
\bottomrule
\end{tabular}
\caption{Ablation on group selection.}
\label{tab:ablation_group}
\end{table}

\section{Limitation}
\label{sec:supp8}

While our proposed method demonstrates promising results and improved efficiency in histopathology VQA, we acknowledge several limitations that outline directions for future research.

\myparagraph{Evaluation on Other Datasets.}
First, our current experimental validation primarily focuses on the TCGA dataset and our in-house private dataset. While this covers a significant amount of variation, the heterogeneity of pathological data across different organs and scanning protocols is vast. To further verify the generalizability of our model, we intend to extend our training and testing to other large-scale public datasets, such as the BCNB (Early Breast Cancer Core-Needle Biopsy) \cite{xu2021predicting} dataset. Evaluating on such diverse cohorts will help ensure our method remains robust across different cancer subtypes and data distributions.

\myparagraph{Lack of Explicit Textual Reasoning.}
Second, while our method offers visual interpretability by highlighting the selected question-relevant patches, it does not currently generate explicit textual explanations justifying \textit{why} these patches were selected. Providing a natural language rationale alongside the final VQA answer would further enhance trust in clinical decision-support systems. We aim to explore the integration of LLMs more deeply in future iterations to bridge this gap between visual attention and semantic reasoning.

\end{document}